\documentclass[runningheads,anonymous]{llncs}
\usepackage[T1]{fontenc}
\usepackage{graphicx}
\usepackage{booktabs}
\usepackage{shadowtext}
\usepackage[misc]{ifsym}
\newcommand{\corr}{(\Letter)}
\usepackage{subfigure}
\usepackage{amsmath}
\usepackage[hidelinks]{hyperref}
\hypersetup{
    colorlinks=true,
    linkcolor=blue,
    citecolor=blue,
    urlcolor=blue
}
\usepackage{url}
\usepackage{multirow}
\usepackage{color}
\usepackage{mwe}
\usepackage{xcolor}
\usepackage{tabularx}
\usepackage{makecell} 
\usepackage{array}
\usepackage{colortbl}
\usepackage{ulem}
\definecolor{myorange}{RGB}{240, 180, 80}
\definecolor{mybrown}{RGB}{120, 70, 40}
\definecolor{mygreen}{RGB}{0,128,0}
\definecolor{mypurple}{RGB}{110,90,160}
\definecolor{mylawcase}{RGB}{127, 142, 186}
\definecolor{mylegal}{RGB}{241, 172, 106}
\definecolor{Color1}{rgb}{1, 0, 0}        
\definecolor{Color2}{rgb}{0.870, 0.435, 0.086} 
\definecolor{Color3}{rgb}{0.949, 0.729, 0.008} 
\definecolor{yellow}{HTML}{F5B82E}
\definecolor{orange}{HTML}{F28D35}
\definecolor{red}{HTML}{E74C3C}
\definecolor{blue}{HTML}{3366CC}
\definecolor{gray}{HTML}{666666}
\definecolor{red1}{RGB}{255, 204, 204} 
\definecolor{red2}{RGB}{255, 153, 153} 
\definecolor{red3}{RGB}{255, 102, 102} 
\definecolor{red4}{RGB}{255, 51, 51}   
\definecolor{red5}{RGB}{255, 0, 0}     
\newcommand{\redhlone}[1]{\sethlcolor{red1}\hl{#1}}
\newcommand{\redhltwo}[1]{\sethlcolor{red2}\hl{#1}}
\newcommand{\redhlthree}[1]{\sethlcolor{red3}\hl{#1}}
\newcommand{\redhlfour}[1]{\sethlcolor{red4}\hl{#1}}
\newcommand{\redhlfive}[1]{\sethlcolor{red5}\hl{#1}}
\usepackage{soul}

\begin{document}

\title{LegalDuet: Learning Fine-grained Representations for Legal Judgment Prediction via a Dual-View Contrastive Learning}

\titlerunning{LegalDuet}


\author{
Buqiang Xu\inst{1}\textsuperscript{$\dagger$} \and 
Xin Dai\inst{1}\textsuperscript{$\dagger$} \and 
Zhenghao Liu\inst{1}\corr \and 
Huiyuan Xie\inst{3} \and 
Xiaoyuan Yi\inst{2} \and 
Shuo Wang\inst{3} \and 
Yukun Yan\inst{3} \and 
Liner Yang\inst{4} \and 
Yu Gu\inst{1} \and 
Ge Yu\inst{1}
}

\makeatletter
\renewcommand{\thefootnote}{}
\footnotetext{\textsuperscript{$\dagger$}Buqiang Xu and Xin Dai contributed equally to this work.}

\renewcommand{\thefootnote}{\arabic{footnote}}
\setcounter{footnote}{0}
\makeatother



\authorrunning{Buqiang Xu et al.}



\institute{
    Northeastern University, Shenyang, Liaoning, China \\
    \email{liuzhenghao@mail.neu.edu.cn}
    \and
    Tsinghua University, Beijing, China \\
    \and
    Microsoft Research Asia, Beijing, China \\
    \and
    Beijing Language and Culture University, Beijing, China \\
}

\maketitle              

\begin{abstract}
Legal Judgment Prediction (LJP) is a fundamental task of legal artificial intelligence, aiming to automatically predict the judgment outcomes of legal cases. Existing LJP models primarily focus on identifying legal triggers within criminal fact descriptions by contrastively training language models.
However, these LJP models overlook the importance of learning to effectively distinguish subtle differences among judgments, which is crucial for producing more accurate predictions. In this paper, we propose LegalDuet, which continuously pretrains language models to learn a more tailored embedding space for representing legal cases. Specifically, LegalDuet designs a dual-view mechanism to continuously pretrain language models: 1) \textbf{Law Case Clustering} retrieves similar cases as hard negatives and employs contrastive training to differentiate among confusing cases; 2) \textbf{Legal Decision Matching} aims to identify legal clues within criminal fact descriptions to align them with the chain of reasoning that contains the correct legal decision.
Our experiments on the CAIL2018 dataset demonstrate the effectiveness of LegalDuet.
Further analysis reveals that LegalDuet improves the ability of pretrained language models to distinguish confusing criminal charges by reducing prediction uncertainty and enhancing the separability of criminal charges. The experiments demonstrate that LegalDuet produces a more concentrated and distinguishable embedding space, effectively aligning criminal facts with corresponding legal decisions.
The code is available at \href{https://github.com/NEUIR/LegalDuet}{https://github.com/NEUIR/LegalDuet}.

\keywords{Legal Judgment Prediction \and Contrastive Learning \and Legal Decision \and Pretrained Language Models}
\end{abstract}

\section{Introduction}
Legal Judgment Prediction (LJP) aims to predict judicial outcomes—including applicable law articles, charges, and imprisonment—based on descriptions of criminal facts. This task plays a vital role in assisting the legal judgment process, improving judicial efficiency, and raising public legal awareness~\cite{xiao2023legal,LegalAI}. Current LJP models predominantly leverage Pretrained Language Models (PLMs), such as BERT~\cite{BERT-2019} and RoBERTa~\cite{liu2019roberta}, to encode criminal fact descriptions and then predict legal judgment labels.

To improve the accuracy and reliability of LJP, recent research has focused on conducting more effective reasoning over criminal facts. Some studies primarily aim to learn fine-grained representations of criminal facts by leveraging handcrafted trigger words and legal attributes~\cite{hu-etal-2018-shot,liu2017predictive,saravanan2009improving} or incorporating predictive information from related LJP tasks~\cite{wu2022towards,yang2019legal,zhong-etal-2018-legal}. With the development of pretraining technologies, the focus has shifted from manually designed features to continuously pretrain language models on legal-domain corpora~\cite{LEGAL-BERT-2020,feng2022legal,shao2020bert,xiao2021lawformer,zhong2019open}, enabling PLMs to capture more semantics from legal texts. However, these models typically employ masked language modeling~\cite{BERT-2019} to help language models learn token-level semantics, rather than learning more tailored representations of entire criminal facts to make more accurate prediction results.

\begin{figure}[t]
    \center
\includegraphics[width=1.0\linewidth]{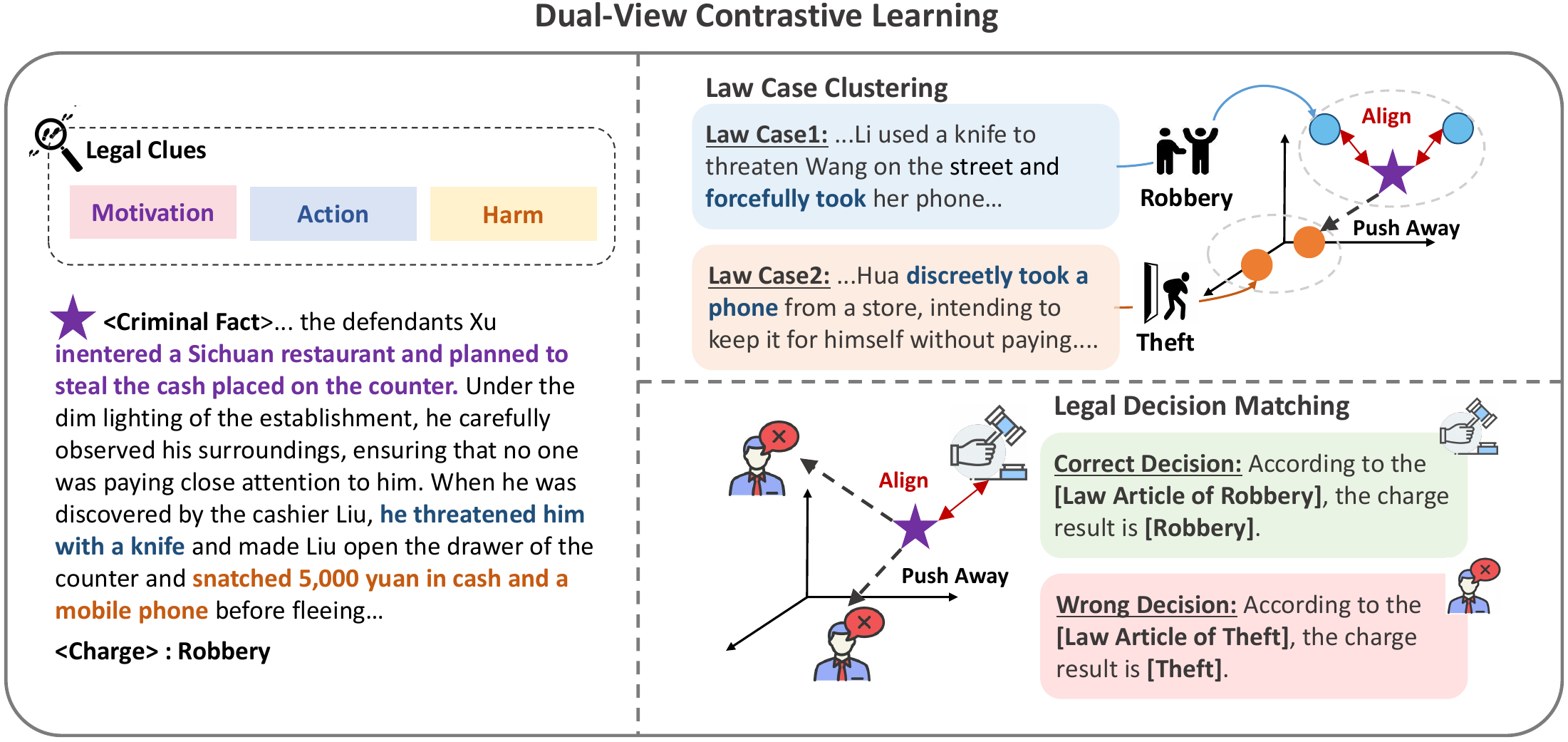}
    \caption{An Example of Dual-View Contrastive Learning Mechanism in LegalDuet. LegalDuet incorporates both Law Case Clustering (LCC) and Legal Decision Matching (LDM) tasks for continuously pretraining language models.}
    \label{fig:1}
\end{figure}

This paper introduces LegalDuet, a framework for continuously pretraining language models to learn more fine-grained representations of legal cases. As illustrated in Fig.~\ref{fig:1}, LegalDuet employs a dual-view contrastive learning mechanism to continuously train language models for legal case representations, which integrates two key tasks: Law Case Clustering and Legal Decision Matching. The Law Case Clustering task leverages hard negatives~\cite{xiong2020approximate} during contrastive learning to enable PLMs to capture subtle differences between similar legal cases. For instance, it helps models discern nuanced differences in the description of legal cases, such as ``forcefully took'' versus ``discreetly took'', which are critical for distinguishing between charges like ``robbery'' and ``theft''.

Besides Law Case Clustering, the Legal Decision Matching task further enhances the representations of legal cases by aligning criminal facts with their corresponding legal decisions. Rather than directly predicting legal judgment labels~\cite{LEGAL-BERT-2020,zhong-etal-2018-legal}, LegalDuet verbalizes, permutes and combines the labels of law articles and charges into reasoning chains to construct alternative legal decisions. PLMs are then tasked with identifying the correct matching between legal cases and decisions. This approach not only encourages PLMs to infer causal relationships between law articles and charges but also stimulates language models to pay more attention to these legal triggers, ultimately improving their ability to represent legal cases comprehensively.

Our experiments on the China Artificial Intelligence and Legal Challenges (CAIL2018)~\cite{CAIL2018} dataset demonstrate the effectiveness of LegalDuet, yielding an improvement on both CAIL-small dataset and CAIL-big dataset compared to the SAILER~\cite{li2023sailer} model. Notably, LegalDuet extends its effectiveness to other PLMs, such as BERT-xs~\cite{zhong2019open} and BERT-Chinese~\cite{BERT-2019}, showcasing its strong generalization capability.
Further analysis reveals that LegalDuet effectively constructs a tailored embedding space for fine-grained legal reasoning by capturing nuanced legal semantics from criminal case facts. LegalDuet produces more compact and coherent representations for legal cases within the same charge category while better separating cases with differing charges. 
These properties of the learned embeddings give LegalDuet the ability to effectively differentiate between similar criminal charges and address ambiguous points in complex criminal cases.


\section{Related Work}
    Early Legal Judgment Prediction (LJP) models primarily rely on feature-based approaches. They aim to extract handcrafted legal cues, such as trigger words and templates pre-defined by legal experts, to support legal judgment predictions~\cite{saravanan2009improving,zeng2005knowledge}. However, the performance of these models is constrained by the quality of these handcrafted features, making it difficult to distinguish between intricate or confusing criminal facts~\cite{gan2021judgment,lauderdale2012supreme}.

In contrast to feature-based approaches, more recent works utilize neural networks to automatically extract legal semantics and clues from criminal facts, such as LSTM~\cite{hochreiter1997long,hu-etal-2018-shot,xu-etal-2020-distinguish,yang2019legal,zhong-etal-2018-legal}, CNN~\cite{johnson2017deep,kim2014convolutional}, and Transformer~\cite{BERT-2019,feng-etal-2022-legal,li2023sailer,yue-2021-neurjudge,zhong2019open}. After encoding legal cases with these neural models, LJP systems can independently predict judgment labels for subtasks like law articles, charges, and imprisonment. To address the gap between different subtasks, some research has focused on modeling task dependencies and employing multi-task learning, which has been shown to improve judgment accuracy~\cite{feng2022legal,ML-LJP,ma2021legal,wu2022towards,yang2019legal,ye2018interpretable,zhong-etal-2018-legal}.

Pretrained Language Models (PLMs), such as BERT~\cite{BERT-2019} and RoBERTa~\cite{liu2019roberta}, have demonstrated strong capabilities in both representing legal cases and making more accurate legal judgments~\cite{zhang2023contrastive}. To bridge the gap between general domain knowledge and the specialized legal domain~\cite{gururangan2020don,shao2020bert}, many studies have focused on continuously training PLMs using legal corpora. Techniques like masked language modeling have been employed to help models learn legal semantics more effectively~\cite{ConSERT-2021,LegalAI}. Additionally, the work related to legal case retrieval~\cite{li2023sailer} pretrains language models to learn the intrinsic logical structures of legal case documents by decoding the reasoning and decisions of judges.
However, these methods mainly focus on token-level~\cite{li2023sailer} clues, often overlooking the need for more nuanced and fine-grained crime representations.

To address this, recent works have focused on two key areas: contrastive training and multi-case reasoning. Contrastive learning encourages the development of building a more tailored embedding space for legal texts, bringing similar cases closer together in this space~\cite{fang2020cert,CLEAR2020,ConSERT-2021}. It has been applied to LJP tasks to enhance the encoding of legal semantics~\cite{gan2023exploiting,ge2021learning,liu2022augmenting,ma2023caseencoder,zhang2023contrastive}. For instance, Zhang et al.~\cite{zhang2023contrastive} propose a supervised contrastive learning approach for LJP, leveraging the structure of law articles to help models differentiate between laws and charges. Similarly, Liu et al.~\cite{liu2022augmenting} incorporate both similar and dissimilar cases during prediction to improve judgment accuracy. Unlike these approaches, LegalDuet focuses on contrastively pretraining PLMs to learn fine-grained representations of legal cases without requiring additional LJP-specific architectures.

Another promising direction is multi-case reasoning, where Graph Neural Networks (GNNs)~\cite{GraphSage,GCN,GAT} are used to propagate semantic information across criminal cases~\cite{KGAT,zhao2020transformer,zhou2019gear}. To enhance legal reasoning in multi-crime scenarios, researchers represent criminal facts as nodes and use predefined legal keywords to form edges between them, constructing hierarchical graphs to enrich case representations~\cite{dong2021legal,feng2022legal,li2021text,yue-2021-neurjudge}. By integrating information from different cases, these graph-based models help distinguish confusing legal texts more effectively~\cite{xu-etal-2020-distinguish}.




\section{Methodology}
In this section, we present our LegalDuet model. We begin by defining the task of Legal Judgment Prediction (LJP) (Sec.~\ref{Sec.3.1}). Next, we describe our dual-view contrastive learning mechanism (Sec.~\ref{Sec.3.2}) to continuously pretrain language models for producing fine-grained representations of criminal facts.

\begin{figure*}[t]
    \centering
    \includegraphics[width=0.9\linewidth]{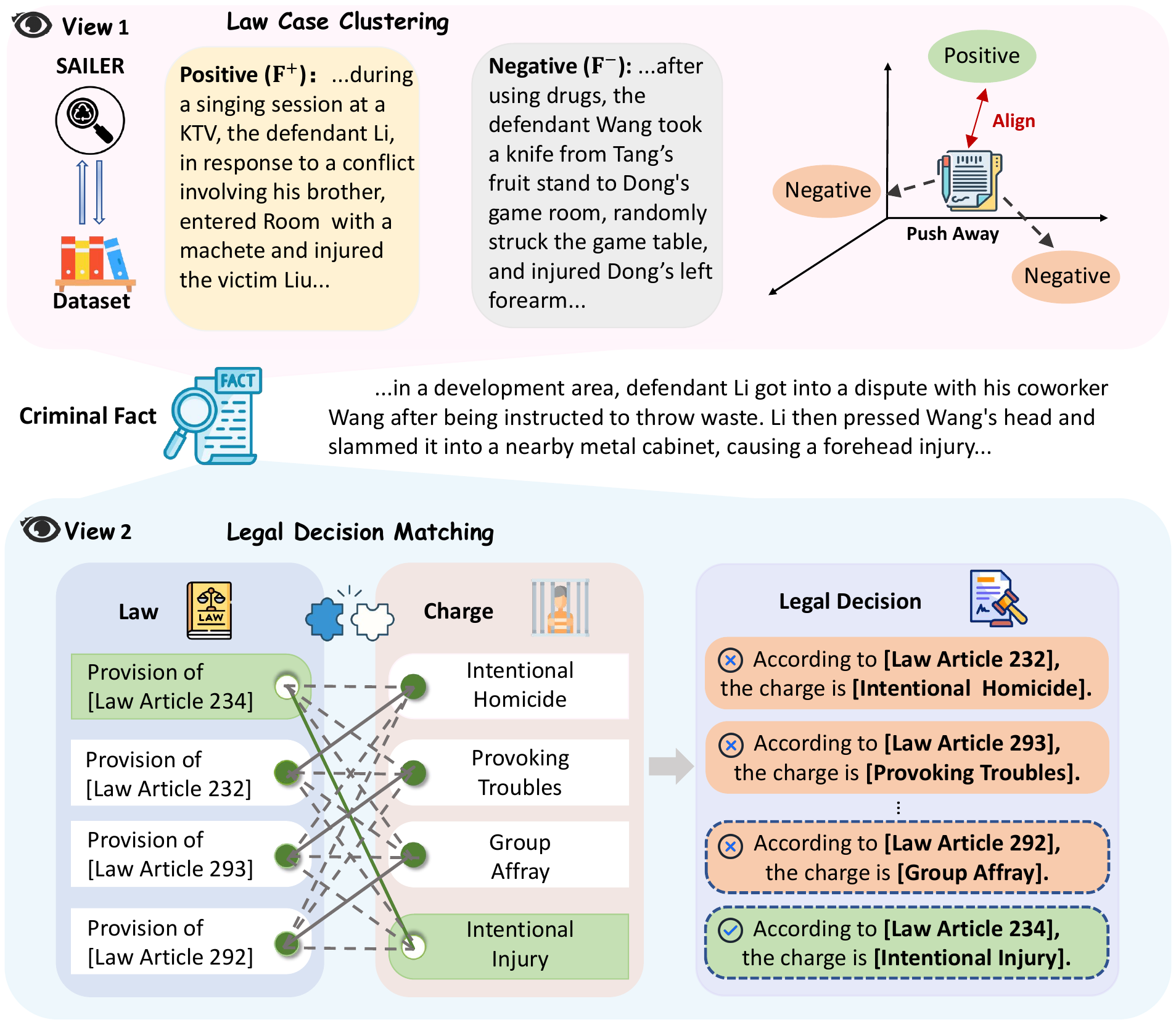}
    \caption{Illustration of Our LegalDuet. 
    }
    \label{fig:2}
\end{figure*}

\subsection{Preliminary of Legal Judgment Prediction}\label{Sec.3.1}
Given a criminal fact $F$, legal judgment prediction (LJP) models first encode its textual description $X_F$ using pretrained language models such as BERT~\cite{BERT-2019}: 
\begin{equation}\label{eq:encode}
h_F = \mathrm{Linear}_F(\mathrm{BERT}(X_F)),
\end{equation}
where $h_F$ is the encoded representation of the criminal fact $F$. The $\mathrm{Linear}_F$ layer reduces the dimensionality of the BERT-generated 768-dimensional embedding to a 256-dimensional space. This reduction helps decrease computational complexity and mitigates the risk of overfitting, consistent with prior studies~\cite{liu2022dimension,zbontar2021barlow}.

For a given LJP task $\mathcal{T}$, the criminal fact representation $h_F$ is then used to predict the label $y_\mathcal{T}$ for that task. LJP tasks aim to identify relevant law articles, charges, or imprisonment terms:
\begin{equation}
P(y_\mathcal{T}|F) = \mathrm{softmax}_{y_\mathcal{T}} (\mathrm{Linear}_{\mathcal{T}}(h_F)),
\end{equation}
where $\mathrm{Linear}_{\mathcal{T}}$ is used to project the hidden representation $h_F$ and compute the logits for different classification labels in the LJP task.

Finally, the LJP models can be trained by using the cross-entropy loss $\mathcal{L}_\mathcal{T}$ for the LJP task $\mathcal{T}$:
\begin{equation}\label{eq:loss}
\mathcal{L}_\mathcal{T}=\mathrm{CrossEntropy}(y^*_\mathcal{T},P(y_\mathcal{T}|F)),
\end{equation}
where $y^*_\mathcal{T}$ is the ground truth label for the given fact $F$.

To better adapt PLMs to legal scenarios, many models focus on continuously pretraining using mask language modeling~\cite{LEGAL-BERT-2020,gururangan2020don}. Similarly, SAILER~\cite{li2023sailer} also employs mask language modeling but proposes a structure-aware approach for legal case retrieval that leverages the inherent structure of legal documents. A complete legal document usually consists of five parts: procedure, criminal fact, reasoning, legal decision, and tail. 
Specifically, SAILER utilizes an asymmetric encoder-decoder architecture to seamlessly integrate multiple pretrained objectives.
In LegalDuet, we utilize SAILER as the backbone to build LJP models.

\subsection{Fine-grained Legal Representation Learning through the Dual-View Contrastive Learning}\label{Sec.3.2}
As shown in Fig.~\ref{fig:2}, LegalDuet leverages a dual-view contrastive learning mechanism to continually pretrain language models from two complementary perspectives: Law Case Clustering (LCC) and the Legal Decision Matching (LDM). Both tasks enable PLMs to conduct more fine-grained representations for criminal facts.
To achieve this, we optimize the encoder models through a multi-task learning approach, using the loss function $\mathcal{L}_{\mathrm{LegalDuet}}$:
\begin{equation}
\mathcal{L}_{\mathrm{LegalDuet}} = \mathcal{L}_{\mathrm{LCC}}+\mathcal{L}_{\mathrm{LDM}},
\end{equation}
where $\mathcal{L}_{\mathrm{LCC}}$ and $\mathcal{L}_{\mathrm{LDM}}$ are the loss functions from the Law Case Clustering task and Legal Decision Matching task. 

Intuitively, Law Case Clustering helps PLMs to learn semantics from legal cases that belong to the same charge and identify subtle distinctions among confusing cases. On the other hand, Legal Decision Matching focuses on aligning criminal facts with relevant legal decisions, guiding the model to map criminal facts to a more appropriate legal decision constructed using law articles and charge details.

\textbf{Law Case Clustering.}
Experienced judges often rely on similar legal cases to guide their judgments, which helps to distinguish subtle differences between confusing cases~\cite{ashley1989modelling,atkinson2005legal}. Building on this, we use Law Case Clustering to assist language models in identifying nuanced distinctions in criminal facts through contrastive training using hard negatives~\cite{xiong2020approximate}.

For a given criminal fact $F$, we contrastively train PLMs using the loss function $\mathcal{L}_{\mathrm{LCC}}$:
\begin{equation}
\mathcal{L}_{\mathrm{LCC}} = -\log \frac{\exp\left(\text{sim}(F, F^{+}) / \tau \right)}{\exp\left(\text{sim}(F, F^{+}) / \tau \right) + \sum_{F^{-} \in \mathcal{N}_F(F)} \exp\left(\text{sim}(F, F^{-}) / \tau \right)},
\end{equation}
where $\tau$ is the temperature used to control the sharpness of the softmax distribution. ${F}^{+}$ is the criminal case that shares the same charge and law article labels with $F$, while ${F}^{-}$ is the negative case that are annotated with different charge or law article labels from $F$. The similarity score between criminal facts $F$ and $F^{+/-}$ is computed using cosine similarity:
\begin{equation}
\text{sim}(F, F^{+/-}) =\cos(h_F, h_{F^{+/-}}).
\end{equation}

To construct the set $\mathcal{N}_F(F)$ of negative legal cases, we use SAILER to retrieve criminal facts that are related to $F$. We select the highest-ranked criminal case that shares the same label as $F$ as the positive fact ${F}^{+}$. For negative samples, we first sample from top-ranked criminal cases with different charges or legal articles. Appendix~\ref{Building Negative Samples} provides a detailed description of our strategy to further process these hard negatives for contrastive training.

\textbf{Legal Decision Matching.}
Professional judges often look for triggers in legal facts to support their judgments~\cite{haar_sawyer_cummings_1977,lauderdale2012supreme,levi2013introduction,sun2023law}. The Legal Decision Matching mechanism builds a decision chain by associating law articles with criminal charges and verbalizing the decision chain. This enables language models to pretrain on identifying judgment triggers inherent in legal facts.

We first define the textual description $X_D$ of legal decisions $D$, which consist of law articles and criminal charges with the following template: 
\begin{equation}\label{eq:template}
{X_D} = \mathrm{Template}(X_{{L}}, X_{{C}}).
\end{equation}
where $X_L$ and $X_C$ are the descriptions of law article $L$ and charge $C$. The decision template is shown in Appendix~\ref{Building Legal Decisions}. 

For a given criminal fact $F$, we contrastively train the PLM to align criminal facts with corresponding legal decisions and distinguish these unmatched legal decisions:
\begin{equation}
\mathcal{L}_{\mathrm{LDM}} = -\log \frac{\exp\left(\text{sim}(F, D^{+}) / \tau \right)}{\exp\left(\text{sim}(F, D^{+}) / \tau \right) + \sum_{D^{-} \in \mathcal{N}_D(F)} \exp\left(\text{sim}(F, D^{-}) / \tau \right)},
\end{equation}
where $\tau$ is the temperature hyperparameter. ${D}^{+}$ represents the legal decision that correctly matches the input criminal fact $F$ according to the ground-truth label, while the negative legal decisions ${D}^{-}$ are those that do not match the input fact $F$, either due to a different charge or law article within the decision.
The legal decision is encoded using the same encoder as the criminal fact:
\begin{equation}
h_{{D}^{+/-}} = \mathrm{Linear}_D(\mathrm{BERT} ({X_{D^{+/-}}})),
\end{equation}
where we use BERT model and $\mathrm{Linear}_D$ for encoding, which shares the same parameters with the encoder in the Eq.~\ref{eq:encode}. 
The similarity of criminal facts and legal decisions can be calculated with cosine similarity:
\begin{equation}
\text{sim}(F, D^{+/-}) =\cos(h_F, h_{D^{+/-}}).
\end{equation}

Rather than randomly selecting negative decisions, we specifically focus on using hard negatives of law articles and charges for contrastive training. Then we incorporate these hard negatives into the candidate pool alongside correct law article and charge. Then we use Eq.~\ref{eq:template} to construct both negative and positive legal decisions.
To construct the set of negative legal decisions $\mathcal{N}_D(F)$, we use the LJP models to predict the the law article and charge labels according to the give fact $F$:
\begin{equation}
(X_{{L}^{-}}, X_{{C}^{-}}) = \text{Classifier}(F).
\end{equation}
The Classifier is initialized with SAILER and then use the training set of CAIL-big to train the LJP models.
After generating these hard negative decisions, we apply the strategy described in Appendix~\ref{Building Negative Samples} for effectively processing these hard negatives for contrastive training to enhance model performance.

\section{Experiment Methodology}
\begin{table}[t]
\small
\caption{Data Statistics of CAIL2018.}
\centering
\begin{tabular}{lrrr} 
\hline
\multirow{2}{*}{\textbf{Datasets}}       & \multicolumn{2}{c}{\textbf{Finetuning}}   & \multicolumn{1}{c}{\textbf{Pretraining}}      \\
\cline{2-4}
& \textbf{Small}       & \textbf{Big}         & \textbf{Rest}  \\ 
\hline
\multicolumn{4}{l}{\textbf{\textit{Raw Dataset}}} \\ 
\hline
Case                 & 142,238              & 1,774,122            & 701,999        \\
Law Articles         & 103                  & 118                  & 59            \\
Charges              & 119                  & 130                  & 62            \\
Term of Penalty      & 11                   & 11                   & 11             \\ 
\hline
\multicolumn{4}{l}{\textbf{\textit{Data Split}}}           \\ 
\hline
Training             & 101,685              & 1,430,135            & 696,999        \\
Development          & 13,787               & 158,759              & 5,000            \\
Test                 & 26,766               & 185,228              & -               \\
\hline
\end{tabular}

\label{tab:1}
\end{table}

In this section, we describe the datasets, baselines, evaluation metrics and implementation details.

\textbf{Datasets.}
We utilize a Chinese LJP dataset CAIL2018~\cite{CAIL2018} in our experiments, which is the same as previous work~\cite{hu-etal-2018-shot,liu2022augmenting,xu-etal-2020-distinguish,yang2019legal,yue-2021-neurjudge,zhong-etal-2018-legal}. It consists of three testing scenarios, CAIL-small, CAIL-big, and CAIL-rest. Specifically, we use CAIL-rest as the pretrained dataset, and both CAIL-small and CAIL-big are adopted during both finetuning and evaluation stages. 
For both finetuning and evaluation, we follow previous work~\cite{xu-etal-2020-distinguish} and keep the same experiment setting\def\thefootnote{1}\footnote{\url{https://github.com/prometheusXN/LADAN/tree/master/data_and_config}}. To guarantee the data quality, we follow Xu et al.~\cite{xu-etal-2020-distinguish} to process the dataset for pretraining and finetuning. We filter out legal cases that contain fewer than 10 meaningful words and are associated with multiple applicable law articles or multiple charges. The legal cases corresponding to law articles and charges with fewer than 100 instances are excluded. The data statistics are shown in Table~\ref{tab:1}.

\textbf{Evaluation Metrics.}
We use the official evaluation metrics of CAIL 2018~\cite{CAIL2018} to assess LJP models, which include Accuracy (Acc), Macro-Precision (MP), and Macro-Recall (MR). These metrics are provided in the official script\def\thefootnote{2}\footnote{\url{https://github.com/china-ai-law-challenge/cail2018}}. In our experiments, we also report the Macro-F1 (F1) score, which is regarded as our primary evaluation metric. The Macro-F1 (F1) score offers a balanced measure between precision and recall.


\textbf{Baselines.}
We utilize several LJP baselines, including feature-based models, neural LJP models, and PLM-based models.

\textit{Feature-based Models.} We use \texttt{TF-IDF}~\cite{salton1988term} to extract legal features and employ \texttt{SVM}~\cite{suykens1999least} to predict the corresponding legal judgments.

\textit{Neural LJP Models.} We compare two convolutional neural network based models, \texttt{TextCNN}~\cite{kim2014convolutional} and \texttt{DPCNN}~\cite{johnson2017deep}, which encode criminal facts using convolutional neural networks. Additionally, six LSTM-based LJP models are evaluated. The \texttt{LSTM}~\cite{hochreiter1997long} model constructs a text classification system using LSTM for predicting legal judgments. LJP models, such as \texttt{TopJudge}~\cite{zhong-etal-2018-legal} and \texttt{MPBFN}~\cite{yang2019legal}, further model the dependencies between different subtasks of LJP. Besides task dependence, many LJP models also focus on fine-grained reasoning by capturing relations among criminal facts. For instance, \texttt{Few-Shot}~\cite{hu-etal-2018-shot} leverages annotated legal attributes to better represent criminal facts, significantly enhancing LJP performance in the few-shot scenario. Models like \texttt{LADAN}~\cite{xu-etal-2020-distinguish} and \texttt{CTM}~\cite{liu2022augmenting} enhance LJP performance by incorporating case relations and frequencies. Specifically, \texttt{LADAN} uses graph neural networks to capture subtle differences among criminal facts, while \texttt{CTM} directly encodes similar and dissimilar facts for LJP.

\textit{PLM-based Models.} For PLM-based methods, we first compare with two BERT-based language models: BERT-Chinese and BERT-xs. These models finetune the \texttt{BERT-Chinese}~\cite{BERT-2019} and \texttt{BERT-xs}~\cite{zhong2019open} models for the LJP tasks. For these multi-tasks based LJP modeling methods, \texttt{NeurJudge}$^+$~\cite{yue-2021-neurjudge} aims to separate fact descriptions into different circumstances using the prediction results of other LJP tasks, aiming to leverage different LJP perspectives to conduct fine-grained prediction. \texttt{SAILER}~\cite{li2023sailer} is a structure-aware pretrained language model for legal case retrieval. It combines both approaches by continuously training BERT to fill in these masked tokens, as well as decode corresponding reasoning results and legal judgments.

\textbf{Implementation Details.}
All experiments are implemented with Pytorch and start from the checkpoints of PLMs from Huggingface Transformers~\cite{vaswani2017attention}.

\textit{Pretraining.} 
We initialize the LJP models using the checkpoint of SAILER\def\thefootnote{3}\footnote{\url{https://huggingface.co/CSHaitao/SAILER_zh}} and continuously train the language model using LegalDuet. 
During pretraining, we set the maximum number of epochs to 5, optimize the parameters using the AdamW optimizer, and configure the learning rate to 1e-5 with a batch size of 32. Following SimCSE~\cite{gao2021simcse}, we adopt the optimal temperature hyperparameter setting ($\tau$ = 0.05) to scale the similarity score.

Then, we present the experimental details for contrastive training.
For Law Case Clustering, we construct a negative sample pool using retrieval results from SAILER and select the top-15 ranked criminal facts as hard negatives for each instance.
For Legal Decision Matching, we employ a fine-tuned SAILER model to classify each criminal fact and identify 3 negative law articles and 3 negative charges to construct the legal decisions. The fine-tuned SAILER model is also trained on the CAIL-big dataset.

\textit{Finetuning.}  
During finetuning, both the classification layer and the BERT parameters are optimized.  
The training process spans 10 epochs, utilizing the AdamW optimizer with a learning rate of 5e-6 and a batch size of 64.  
Further implementation details of the baselines are provided in Appendix~\ref{implementation}.

\section{Evaluation Results}
\begin{table}[!t]
\centering
\caption{LJP Performance on the CAIL-big Dataset. 
The best results are in \textbf{bold}, and the \uline{underlined} scores indicate the second-best results.}
\resizebox{\linewidth}{!}{ 
\begin{tabular}{lcccccccccccc} 
\toprule
\multirow{2}{*}{\textbf{Model}} & \multicolumn{4}{c}{\textbf{Law Articles}} & \multicolumn{4}{c}{\textbf{Charges}} & \multicolumn{4}{c}{\textbf{Imprisonment}} \\
\cmidrule(lr){2-5} \cmidrule(lr){6-9} \cmidrule(lr){10-13}
& Acc & MP & MR & F1 & Acc & MP & MR & F1 & Acc & MP & MR & F1 \\
\midrule
TF-IDF+SVM               & 94.83 & 83.99 & 68.21 & 72.23 & 94.60 & 86.71 & 71.18 & 75.36 & 51.63 & 39.62 & 29.41 & 30.28 \\
TextCNN                 & 93.53 & 74.97 & 59.40 & 62.37 & 93.10 & 78.44 & 62.12 & 65.41 & 51.15 & 39.76 & 28.23 & 28.07 \\
DPCNN                   & 93.81 & 76.53 & 65.91 & 68.57 & 93.58 & 78.57 & 68.83 & 71.40 & 51.65 & 40.56 & 30.81 & 33.57 \\
LSTM                    & 94.27 & 76.69 & 63.78 & 65.73 & 93.87 & 78.58 & 66.67 & 68.81 & 52.29 & 36.67 & 32.41 & 32.71 \\
TopJudge                & 94.54 & 76.83 & 66.98 & 68.82 & 94.18 & 81.08 & 69.44 & 71.16 & 52.64 & 38.86 & 34.19 & 33.71 \\
MPBFN                   & 95.85 & 85.08 & 72.99 & 76.24 & 95.51 & 88.81 & 74.81 & 78.82 & 56.59 & 49.55 & 38.28 & 40.20 \\
LADAN                   & 96.34 & 85.75 & 78.45 & 80.88 & 96.22 & 87.64 & 81.60 & 83.74 & 58.60 & 50.39 & 43.21 & 44.67 \\
CTM                     & 97.24 & 86.56 & 81.45 & 82.11 & 97.00 & 87.45 & 83.53 & 83.73 & 56.70 & 46.81 & 39.81 & 41.65 \\
Few-Shot                & 96.48 & 86.78 & 78.30 & 80.98 & 96.41 & 90.13 & 82.59 & 85.34 & 58.26 & 51.71 & 41.53 & 44.31 \\
NeurJudge$^{+}$         & 94.88 & 85.01 & 75.68 & 78.12 & 95.47 & 81.77 & 72.77 & 75.00 & 56.93 & 47.17 & 40.72 & 42.10 \\
\midrule
BERT-xs\cellcolor{mygreen!5}                 & 96.20\cellcolor{mygreen!5} & 82.30\cellcolor{mygreen!5} & 75.29\cellcolor{mygreen!5} & 77.29\cellcolor{mygreen!5} & 96.09\cellcolor{mygreen!5} & 87.15\cellcolor{mygreen!5} & 79.64\cellcolor{mygreen!5} & 81.72\cellcolor{mygreen!5} & 58.06\cellcolor{mygreen!5} & 48.07\cellcolor{mygreen!5} & 43.47\cellcolor{mygreen!5} & 44.71\cellcolor{mygreen!5} \\
w/ LegalDuet\cellcolor{mygreen!5}            & 96.76\cellcolor{mygreen!5} & 85.55\cellcolor{mygreen!5} & 79.31\cellcolor{mygreen!5} & 80.94\cellcolor{mygreen!5} & 96.70\cellcolor{mygreen!5} & 89.52\cellcolor{mygreen!5} & 83.35\cellcolor{mygreen!5} & 85.42\cellcolor{mygreen!5} & 58.89\cellcolor{mygreen!5} & 49.00\cellcolor{mygreen!5} & 45.31\cellcolor{mygreen!5} & 46.24\cellcolor{mygreen!5} \\
\midrule
BERT-Chinese\cellcolor{pink!20}            & 97.09\cellcolor{pink!20} & 87.65\cellcolor{pink!20} & 82.28\cellcolor{pink!20} & 83.90\cellcolor{pink!20} & 97.05\cellcolor{pink!20} & 90.89\cellcolor{pink!20} & 85.93\cellcolor{pink!20} & 87.65\cellcolor{pink!20} & 62.17\cellcolor{pink!20} & \uline{54.45}\cellcolor{pink!20} & 49.82\cellcolor{pink!20} & 51.34\cellcolor{pink!20} \\
w/ LegalDuet\cellcolor{pink!20}            & 97.29\cellcolor{pink!20} & \textbf{88.76}\cellcolor{pink!20} & \uline{84.52}\cellcolor{pink!20} & \uline{85.93}\cellcolor{pink!20} & 97.27\cellcolor{pink!20} & 91.51\cellcolor{pink!20} & 87.73\cellcolor{pink!20} & 89.14\cellcolor{pink!20} & 62.21\cellcolor{pink!20} & 53.80\cellcolor{pink!20} & 51.40\cellcolor{pink!20} & 51.79\cellcolor{pink!20} \\
\midrule
SAILER\cellcolor{blue!8}                  & \uline{97.35}\cellcolor{blue!8} & 88.69\cellcolor{blue!8} & 84.29\cellcolor{blue!8} & 85.89\cellcolor{blue!8} & \uline{97.34}\cellcolor{blue!8} & \textbf{91.88}\cellcolor{blue!8} & \uline{87.94}\cellcolor{blue!8} & \uline{89.49}\cellcolor{blue!8} & \uline{62.75}\cellcolor{blue!8} & 54.15\cellcolor{blue!8} & \textbf{52.39}\cellcolor{blue!8} & \uline{52.77}\cellcolor{blue!8} \\
w/ LegalDuet\cellcolor{blue!8}            & \textbf{97.39}\cellcolor{blue!8} & \uline{88.73}\cellcolor{blue!8} & \textbf{85.20}\cellcolor{blue!8} & \textbf{86.32}\cellcolor{blue!8} & \textbf{97.44}\cellcolor{blue!8} & \uline{91.86}\cellcolor{blue!8} & \textbf{88.80}\cellcolor{blue!8} & \textbf{89.97}\cellcolor{blue!8} & \textbf{63.06}\cellcolor{blue!8} & \textbf{55.05}\cellcolor{blue!8} & \uline{52.34}\cellcolor{blue!8} & \textbf{53.16}\cellcolor{blue!8} \\
\bottomrule
\end{tabular}
}
\label{CAIL-big}
\end{table}

In this section, we evaluate the effectiveness of LegalDuet on LJP tasks. We first present its overall performance and conduct ablation studies to assess the contributions of each module of LegalDuet. Subsequently, we conduct detailed analyses to explore the characteristics of LegalDuet. Additionally, we visualize attention weights and show case studies in Appendices~\ref{attention} and~\ref{case}, respectively.

\begin{table}[!t]
\centering
\caption{LJP Performance on the CAIL-small Dataset. 
The best results are in \textbf{bold}, and the \uline{underlined} scores indicate the second-best results.}
\resizebox{\linewidth}{!}{ 
\begin{tabular}{lcccccccccccc} 
\toprule
\multirow{2}{*}{\textbf{Model}} & \multicolumn{4}{c}{\textbf{Law Articles}} & \multicolumn{4}{c}{\textbf{Charges}} & \multicolumn{4}{c}{\textbf{Imprisonment}} \\
\cmidrule(lr){2-5} \cmidrule(lr){6-9} \cmidrule(lr){10-13}
& Acc & MP & MR & F1 & Acc & MP & MR & F1 & Acc & MP & MR & F1 \\
\midrule
TF-IDF+SVM & 77.76 & 77.89 & 72.84 & 72.70 & 80.31 & 82.74 & 77.56 & 78.30 & 35.33 & 29.24 & 26.74 & 26.13 \\
TEXTCNN    & 75.81 & 71.33 & 70.40 & 68.37 & 76.90 & 74.80 & 74.95 & 73.05 & 34.31 & 32.84 & 30.10 & 28.93 \\
DPCNN      & 76.10 & 70.93 & 70.46 & 68.89 & 79.95 & 76.01 & 77.75 & 76.23 & 35.03 & 31.31 & 29.51 & 29.78 \\
LSTM       & 77.79 & 76.07 & 73.67 & 72.55 & 81.60 & 80.22 & 79.16 & 78.56 & 36.53 & 26.91 & 28.82 & 26.41 \\
TopJudge   & 77.29 & 76.47 & 74.67 & 73.08 & 82.91 & 80.86 & 79.75 & 79.03 & 36.24 & 28.52 & 29.90 & 27.50 \\
MPBFN      & 79.88 & 79.36 & 75.50 & 75.29 & 83.02 & 84.06 & 80.93 & 81.41 & 37.04 & 36.87 & 29.81 & 29.05 \\
LADAN      & 80.92 & 77.74 & 78.79 & 77.14 & 84.97 & 83.34 & 83.67 & 83.03 & 37.70 & 37.04 & 33.48 & 34.94 \\
CTM        & 83.10 & 79.12 & 81.81 & 79.23 & 87.92 & 86.31 & 86.08 & 85.71 & 37.81 & 34.66 & 30.32 & 30.90 \\
Few-Shot   & 79.17 & 77.54 & 74.78 & 74.00 & 81.91 & 82.63 & 79.61 & 79.64 & 35.87 & 34.38 & 28.86 & 30.08 \\
NeurJudge+ & 80.02 & 77.39 & 76.97 & 75.50 & 83.24 & 83.26 & 80.95 & 83.24 & 38.59 & 35.35 & 33.19 & 32.33 \\
\midrule
BERT-xs\cellcolor{mygreen!5}       
& 80.99\cellcolor{mygreen!5} 
& 79.23\cellcolor{mygreen!5}  
& 76.62\cellcolor{mygreen!5} 
& 76.03\cellcolor{mygreen!5}  
& 85.08\cellcolor{mygreen!5}  
& 83.68\cellcolor{mygreen!5} 
& 82.45\cellcolor{mygreen!5}  
& 82.28\cellcolor{mygreen!5} 
& 38.59\cellcolor{mygreen!5} 
& 32.77\cellcolor{mygreen!5}  
& 33.15\cellcolor{mygreen!5} 
& 31.45\cellcolor{mygreen!5}  \\
w/ LegalDuet\cellcolor{mygreen!5} 
& 83.20\cellcolor{mygreen!5} 
& 80.54\cellcolor{mygreen!5} 
& 80.23\cellcolor{mygreen!5} 
& 79.21\cellcolor{mygreen!5} 
& 87.48\cellcolor{mygreen!5} 
& 85.47\cellcolor{mygreen!5} 
& 83.60\cellcolor{mygreen!5} 
& 83.85\cellcolor{mygreen!5} 
& 39.20\cellcolor{mygreen!5} 
& 35.97\cellcolor{mygreen!5} 
& 33.72\cellcolor{mygreen!5} 
& 32.65\cellcolor{mygreen!5}  \\
\midrule
BERT-Chinese\cellcolor{pink!20}
& 82.47\cellcolor{pink!20}
& 79.75\cellcolor{pink!20}
& 79.15\cellcolor{pink!20}
& 78.51\cellcolor{pink!20}
& 87.66\cellcolor{pink!20}
& 86.36\cellcolor{pink!20}
& 85.94\cellcolor{pink!20}
& 85.65\cellcolor{pink!20}
& 41.93\cellcolor{pink!20}
& 39.21\cellcolor{pink!20}
& 37.53\cellcolor{pink!20}
& 36.85\cellcolor{pink!20} \\
w/ LegalDuet\cellcolor{pink!20}
& 83.93\cellcolor{pink!20}
& \uline{82.48}\cellcolor{pink!20}
& 81.86\cellcolor{pink!20}
& \uline{81.00}\cellcolor{pink!20}
& 89.74\cellcolor{pink!20} 
& 87.92\cellcolor{pink!20}
& 87.27\cellcolor{pink!20}
& 87.19\cellcolor{pink!20}
& 42.43\cellcolor{pink!20}
& \uline{39.61}\cellcolor{pink!20}
& \uline{38.08}\cellcolor{pink!20}
& \textbf{37.91}\cellcolor{pink!20} \\
\midrule
SAILER\cellcolor{blue!8}
& \uline{84.23}\cellcolor{blue!8}
& 81.67\cellcolor{blue!8}
& \uline{82.59}\cellcolor{blue!8}
& 81.00 \cellcolor{blue!8}
& \uline{89.75}\cellcolor{blue!8}
& \uline{88.12}\cellcolor{blue!8}
& \uline{87.65}\cellcolor{blue!8}
& \uline{87.59}\cellcolor{blue!8}
& \uline{42.76}\cellcolor{blue!8}
& 38.87\cellcolor{blue!8}
& 38.00\cellcolor{blue!8}
& 37.15\cellcolor{blue!8} \\
w/ LegalDuet\cellcolor{blue!8}     & \textbf{85.90}\cellcolor{blue!8} & \textbf{83.92}\cellcolor{blue!8} & \textbf{83.26}\cellcolor{blue!8} & \textbf{82.65}\cellcolor{blue!8} & \textbf{90.47}\cellcolor{blue!8} & \textbf{88.90}\cellcolor{blue!8} & \textbf{88.36}\cellcolor{blue!8} & \textbf{88.29}\cellcolor{blue!8} & \textbf{43.25}\cellcolor{blue!8} & \textbf{40.62}\cellcolor{blue!8} & \textbf{38.25}\cellcolor{blue!8} & \uline{37.85}\cellcolor{blue!8} \\
\bottomrule
\end{tabular}
}
\label{CAIL-small}
\end{table}

\subsection{Overall Performance}

The LJP performance of LegalDuet and the baseline models is presented in Table~\ref{CAIL-big} and Table~\ref{CAIL-small}. LegalDuet demonstrates its effectiveness by outperforming all baseline models across various testing scenarios, which thrives on its ability to learn more fine-grained representations of criminal facts.

As shown in Table~\ref{CAIL-big}, LegalDuet demonstrates its effectiveness by surpassing SAILER, particularly in the law articles prediction task. Furthermore, as presented in Table~\ref{CAIL-small}, the performance gains of LegalDuet become even more pronounced, highlighting its strong capability in few-shot learning scenarios. Notably, our approach eliminates the need for complex model architectures, such as explicitly modeling dependencies among different LJP tasks or extracting fine-grained legal cues. Despite leveraging only law articles and charges during continuous pretraining, LegalDuet achieves substantial improvements across other tasks, including imprisonment prediction, demonstrating its adaptability to various legal judgment prediction tasks. Additionally, LegalDuet exhibits strong generalization by extending its effectiveness to different PLMs, including BERT-xs and BERT-Chinese.




\begin{table}[!t]
\centering
\caption{Performance of Ablation Models on LJP Tasks. All models are evaluated on the CAIL-small dataset. LDM and LCC indicate the Law Decision Matching and Law Case Clustering tasks, respectively.}
\resizebox{\linewidth}{!}{ 
\begin{tabular}{lcccccccccccc} 
\toprule
\multirow{2}{*}{\textbf{Model}} & \multicolumn{4}{c}{\textbf{Law Articles}} & \multicolumn{4}{c}{\textbf{Charges}} & \multicolumn{4}{c}{\textbf{Imprisonment}} \\
\cmidrule(lr){2-5} \cmidrule(lr){6-9} \cmidrule(lr){10-13}
& Acc & MP & MR & F1 & Acc & MP & MR & F1 & Acc & MP & MR & F1 \\
\midrule
BERT-xs\cellcolor{mygreen!5} & 80.99\cellcolor{mygreen!5} & 79.23\cellcolor{mygreen!5} & 76.62\cellcolor{mygreen!5} & 76.03\cellcolor{mygreen!5} & 85.08\cellcolor{mygreen!5} & 83.68\cellcolor{mygreen!5} & 82.45\cellcolor{mygreen!5} & 82.28\cellcolor{mygreen!5} & 38.59\cellcolor{mygreen!5} & 32.77\cellcolor{mygreen!5} & 33.15\cellcolor{mygreen!5} & 31.45\cellcolor{mygreen!5} \\
w/ LDM\cellcolor{mygreen!5} & 81.96\cellcolor{mygreen!5} & 79.76\cellcolor{mygreen!5} & \uline{79.30}\cellcolor{mygreen!5} & 78.10\cellcolor{mygreen!5} & 86.49\cellcolor{mygreen!5} & 84.73\cellcolor{mygreen!5} & 82.86\cellcolor{mygreen!5} & 82.99\cellcolor{mygreen!5} & \textbf{39.40}\cellcolor{mygreen!5} & 35.13\cellcolor{mygreen!5} & \textbf{34.02}\cellcolor{mygreen!5} & \textbf{32.72}\cellcolor{mygreen!5} \\
w/ LCC\cellcolor{mygreen!5} & \uline{83.07}\cellcolor{mygreen!5} & \uline{80.34}\cellcolor{mygreen!5} & 79.23\cellcolor{mygreen!5} & \uline{78.42}\cellcolor{mygreen!5} & \uline{86.86}\cellcolor{mygreen!5} & \uline{85.17}\cellcolor{mygreen!5} & \uline{83.39}\cellcolor{mygreen!5} & \uline{83.63}\cellcolor{mygreen!5} & \uline{39.38}\cellcolor{mygreen!5} & \uline{35.53}\cellcolor{mygreen!5} & \uline{33.84}\cellcolor{mygreen!5} & 32.41\cellcolor{mygreen!5} \\
LegalDuet\cellcolor{mygreen!5}      & \textbf{83.20}\cellcolor{mygreen!5} & \textbf{80.54}\cellcolor{mygreen!5} & \textbf{80.23}\cellcolor{mygreen!5} & \textbf{79.21}\cellcolor{mygreen!5} & \textbf{87.48}\cellcolor{mygreen!5} & \textbf{85.47}\cellcolor{mygreen!5} & \textbf{83.60}\cellcolor{mygreen!5} & \textbf{83.85}\cellcolor{mygreen!5} & 39.20\cellcolor{mygreen!5} & \textbf{35.97}\cellcolor{mygreen!5} & 33.72\cellcolor{mygreen!5} & \uline{32.65}\cellcolor{mygreen!5} \\
\midrule
BERT-Chinese\cellcolor{pink!20} & 82.47\cellcolor{pink!20} & 79.75\cellcolor{pink!20} & 79.15\cellcolor{pink!20} & 78.51\cellcolor{pink!20} & 87.66\cellcolor{pink!20} & 86.36\cellcolor{pink!20} & 85.94\cellcolor{pink!20} & 85.65\cellcolor{pink!20} & 41.93\cellcolor{pink!20} & 39.21\cellcolor{pink!20} & 37.53\cellcolor{pink!20} & 36.85\cellcolor{pink!20} \\
w/ LDM\cellcolor{pink!20}& \textbf{84.22}\cellcolor{pink!20} & 81.69\cellcolor{pink!20} & \uline{81.95}\cellcolor{pink!20} & \uline{81.00}\cellcolor{pink!20} & \uline{89.17}\cellcolor{pink!20} & \uline{87.27}\cellcolor{pink!20} & 86.42\cellcolor{pink!20} & \uline{86.47}\cellcolor{pink!20} & 41.93\cellcolor{pink!20} & \textbf{39.86}\cellcolor{pink!20} & \textbf{38.47}\cellcolor{pink!20} & \uline{37.66}\cellcolor{pink!20} \\
w/ LCC\cellcolor{pink!20} & 83.66\cellcolor{pink!20} & \uline{81.99}\cellcolor{pink!20} & \textbf{82.00}\cellcolor{pink!20} & 80.86\cellcolor{pink!20} & 89.10\cellcolor{pink!20} & 86.93\cellcolor{pink!20} & \uline{86.64}\cellcolor{pink!20} & 86.35\cellcolor{pink!20} & \uline{42.20}\cellcolor{pink!20} & 39.33\cellcolor{pink!20} & 37.91\cellcolor{pink!20} & 37.29\cellcolor{pink!20} \\
LegalDuet\cellcolor{pink!20} & \uline{83.93}\cellcolor{pink!20} & \textbf{82.48}\cellcolor{pink!20} & 81.86\cellcolor{pink!20} & \textbf{81.00}\cellcolor{pink!20} & \textbf{89.74}\cellcolor{pink!20} & \textbf{87.92}\cellcolor{pink!20} & \textbf{87.27}\cellcolor{pink!20} & \textbf{87.19}\cellcolor{pink!20} & \textbf{42.43}\cellcolor{pink!20} & \uline{39.61}\cellcolor{pink!20} & \uline{38.08}\cellcolor{pink!20} & \textbf{37.91}\cellcolor{pink!20} \\
\midrule
SAILER\cellcolor{blue!8}
& 84.23\cellcolor{blue!8}
& 81.67\cellcolor{blue!8}
& 82.59\cellcolor{blue!8}
& 81.00\cellcolor{blue!8}
& 89.75\cellcolor{blue!8}
& 88.12\cellcolor{blue!8}
& 87.65\cellcolor{blue!8}
& 87.59\cellcolor{blue!8}
& 42.76\cellcolor{blue!8}
& 38.87\cellcolor{blue!8}
& \uline{38.00}\cellcolor{blue!8}
& 37.15\cellcolor{blue!8}\\
w/ LDM\cellcolor{blue!8}
& \uline{85.68}\cellcolor{blue!8}
& 83.32\cellcolor{blue!8}
& \textbf{83.32}\cellcolor{blue!8}
& \uline{82.45}\cellcolor{blue!8}
& 89.86\cellcolor{blue!8}
& 88.56\cellcolor{blue!8}
& \textbf{88.46}\cellcolor{blue!8}
& \uline{88.16}\cellcolor{blue!8}
& \textbf{43.58}\cellcolor{blue!8}
& \textbf{41.61}\cellcolor{blue!8}
& 37.44\cellcolor{blue!8}
& \uline{37.63}\cellcolor{blue!8} \\
w/ LCC\cellcolor{blue!8}
& 84.83\cellcolor{blue!8}
& \uline{83.76}\cellcolor{blue!8} 
& 83.12\cellcolor{blue!8}
& 82.30\cellcolor{blue!8}
& \uline{90.21}\cellcolor{blue!8}
& \uline{88.80}\cellcolor{blue!8}
& 88.03\cellcolor{blue!8}
& 87.97\cellcolor{blue!8}
& 42.95\cellcolor{blue!8}
& 39.95\cellcolor{blue!8}
& 37.76\cellcolor{blue!8}
& 37.46\cellcolor{blue!8} \\
LegalDuet\cellcolor{blue!8} & \textbf{85.90}\cellcolor{blue!8} & \textbf{83.92}\cellcolor{blue!8} & \uline{83.26}\cellcolor{blue!8} & \textbf{82.65}\cellcolor{blue!8} & \textbf{90.47}\cellcolor{blue!8} & \textbf{88.90}\cellcolor{blue!8} & \uline{88.36}\cellcolor{blue!8} & \textbf{88.29}\cellcolor{blue!8} & \uline{43.25}\cellcolor{blue!8} & \uline{40.62}\cellcolor{blue!8} & \textbf{38.25}\cellcolor{blue!8} & \textbf{37.85}\cellcolor{blue!8} \\
\bottomrule
\end{tabular}
}
\label{ablation}
\end{table}

\subsection{Ablation Study}
Then, we conduct ablation studies, as presented in Table~\ref{ablation}, to evaluate the contributions of different modules in LegalDuet across Law Articles, Charges, and Imprisonment prediction tasks.

Compared to vanilla PLMs, the LCC module significantly enhances LJP performance across various tasks, demonstrating its effectiveness. This highlights the capability of our Law Case Clustering module in extracting case-specific legal cues—such as critical actions, motivations, and contextual details—that are essential for fine-grained legal reasoning.
Similarly, the LDM module also surpasses vanilla PLMs, with its impact being particularly evident in tasks requiring precise alignment, such as Law Articles and Imprisonment prediction. This underscores the effectiveness of the Legal Decision Matching module in aligning criminal facts with corresponding legal decisions, ensuring that the model integrates legal knowledge into its reasoning process.
When both modules are incorporated, LegalDuet consistently outperforms the baseline across all tasks, emphasizing their complementary roles. Law Case Clustering aids in distinguishing between criminal facts linked to different charges by extracting case-specific clues, while Legal Decision Matching ensures alignment between criminal facts and legal decisions. Together, these modules empower LegalDuet to capture fine-grained semantics and extract legally relevant information, thereby enhancing the model's overall LJP capabilities.

\begin{figure}[t!]
    \centering
    \subfigure[Entropy Distribution of Charge Predictions.\label{fig:3a}]{%
        \includegraphics[width=0.32\textwidth]{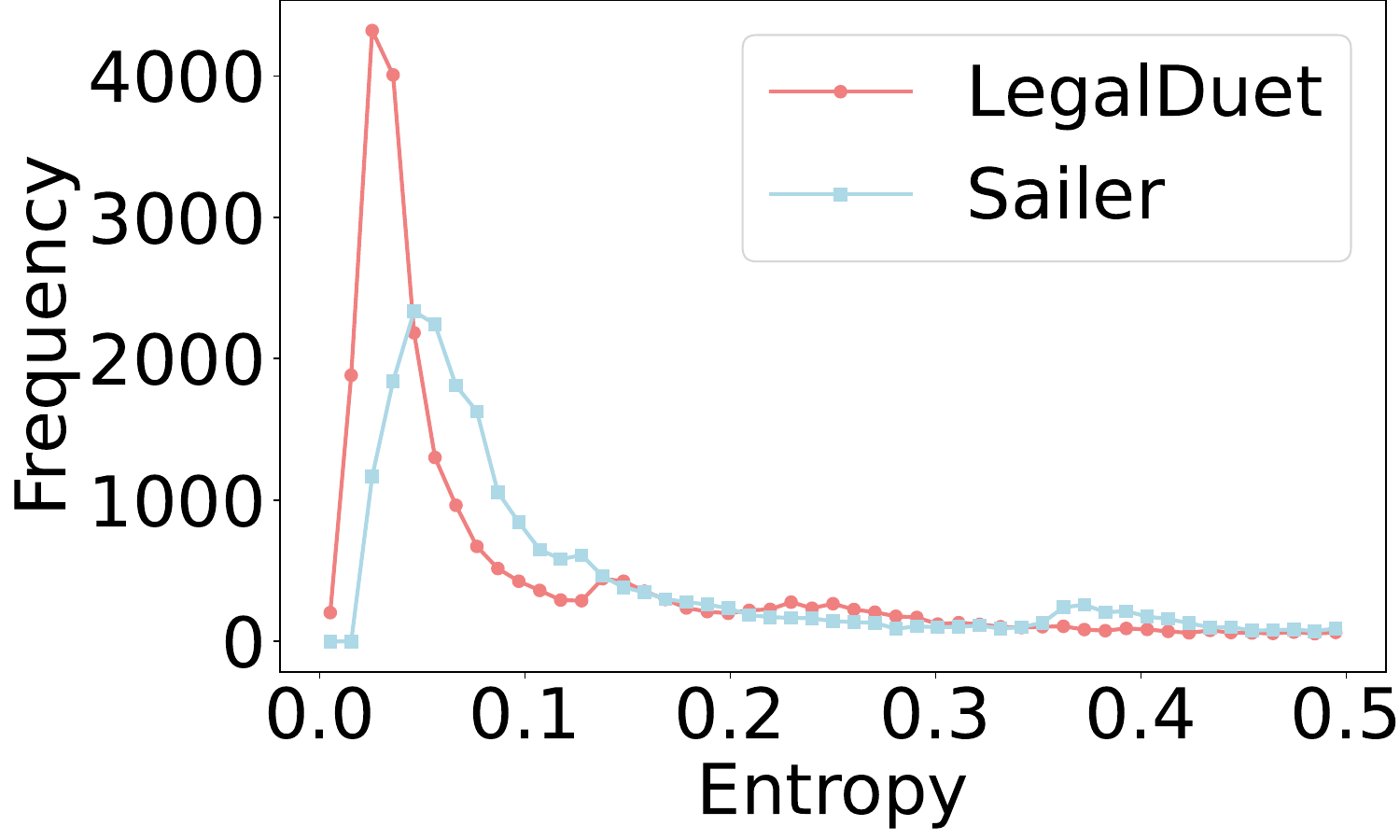}%
    }
    \hfill
    \subfigure[Entropy Distribution of Law Articles Predictions.\label{fig:3b}]{%
        \includegraphics[width=0.32\textwidth]{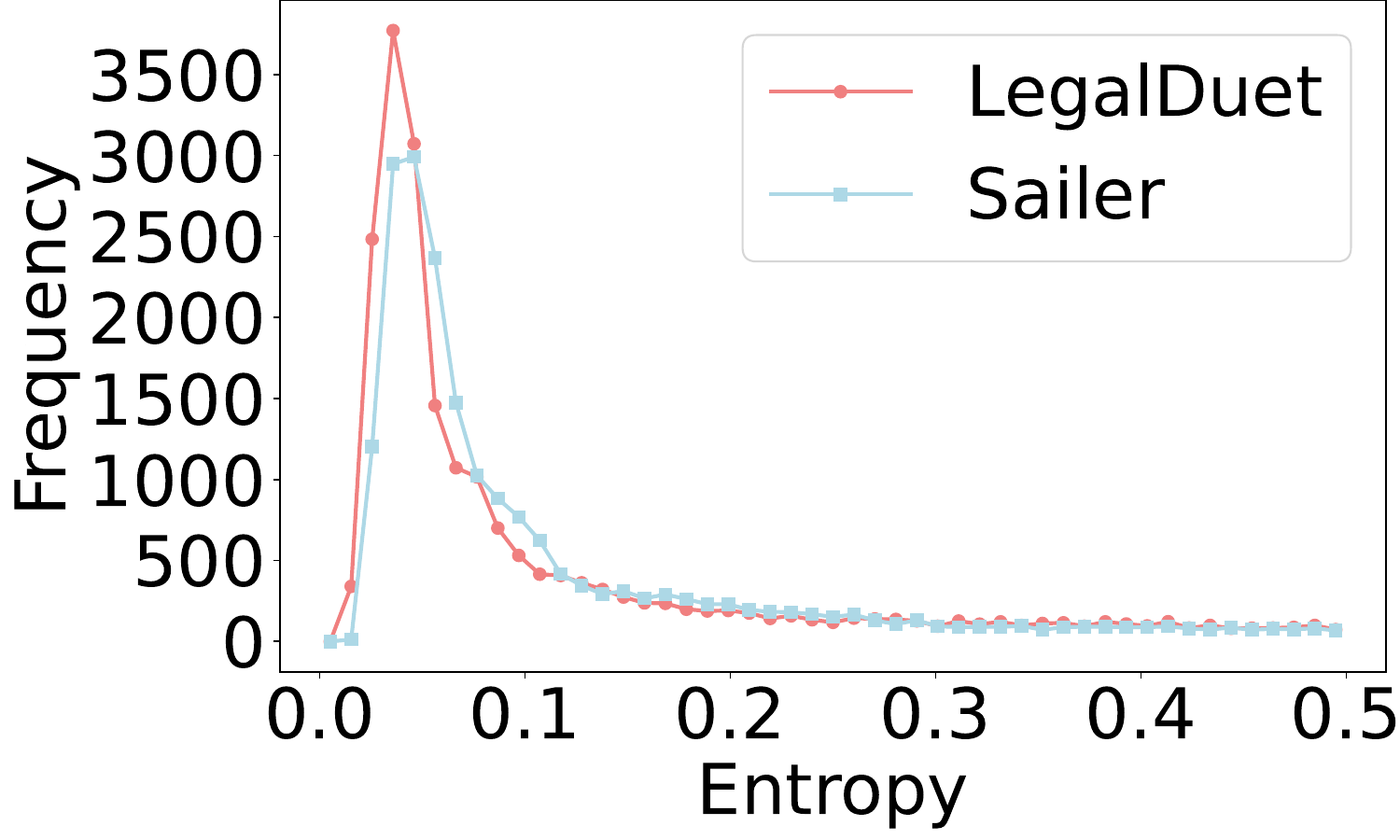}%
    }
    \hfill
    \subfigure[Entropy Distribution of Imprisonment Predictions.\label{fig:3c}]{%
        \includegraphics[width=0.32\textwidth]{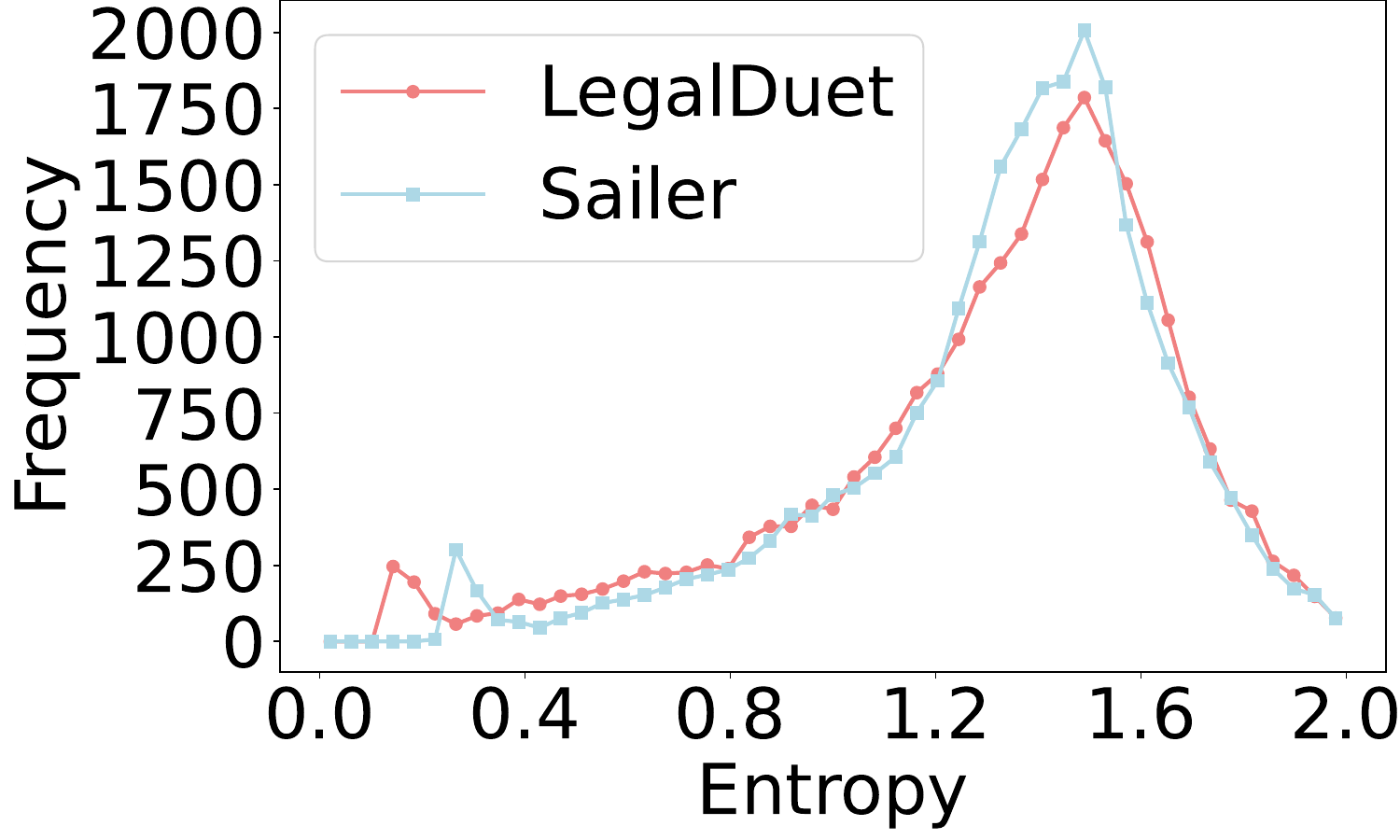}%
    }
    \hfill
    
    \caption{Entropy Distributions of Legal Judgment Predictions.}
    \label{fig:3}
\end{figure}

\subsection{Learned Embeddings of Criminal Facts Optimized by LegalDuet}
In this subsection, we evaluate the effectiveness of LegalDuet in learning fine-grained representations of criminal facts through two sets of experiments. The first experiment examines model uncertainty in legal judgment prediction (LJP) across different models, while the second investigates the learned embedding space to assess the impact of LegalDuet based pretraining.

\textbf{Prediction Confidence of Different LJP Models.}
We begin by quantitatively analyzing model uncertainty in LJP by presenting cross entropy scores, as illustrated in Fig.~\ref{fig:3}. A higher cross-entropy score indicates lower model confidence in predicting the ground truth.

As shown in the evaluation results, LegalDuet generally reduces prediction entropy scores compared to SAILER, enabling LJP models to make more confident and precise predictions. Across different testing scenarios, LegalDuet is particularly effective in lowering entropy scores for the charge prediction task, illustrating that it facilitates better semantic alignment between criminal facts and charges. This improvement may stem from LegalDuet's approach of verbalizing charge decisions and leveraging contrastive training to distinguish them, thereby enhancing charge comprehension and representation learning ability of PLMs. Additional experimental results for other PLMs are provided in Appendix~\ref{Entropy}.

\begin{figure}[t]
    \centering
    \subfigure[DBI Reduction Values.\label{fig:4a}]{%
        \includegraphics[width=0.32\textwidth]{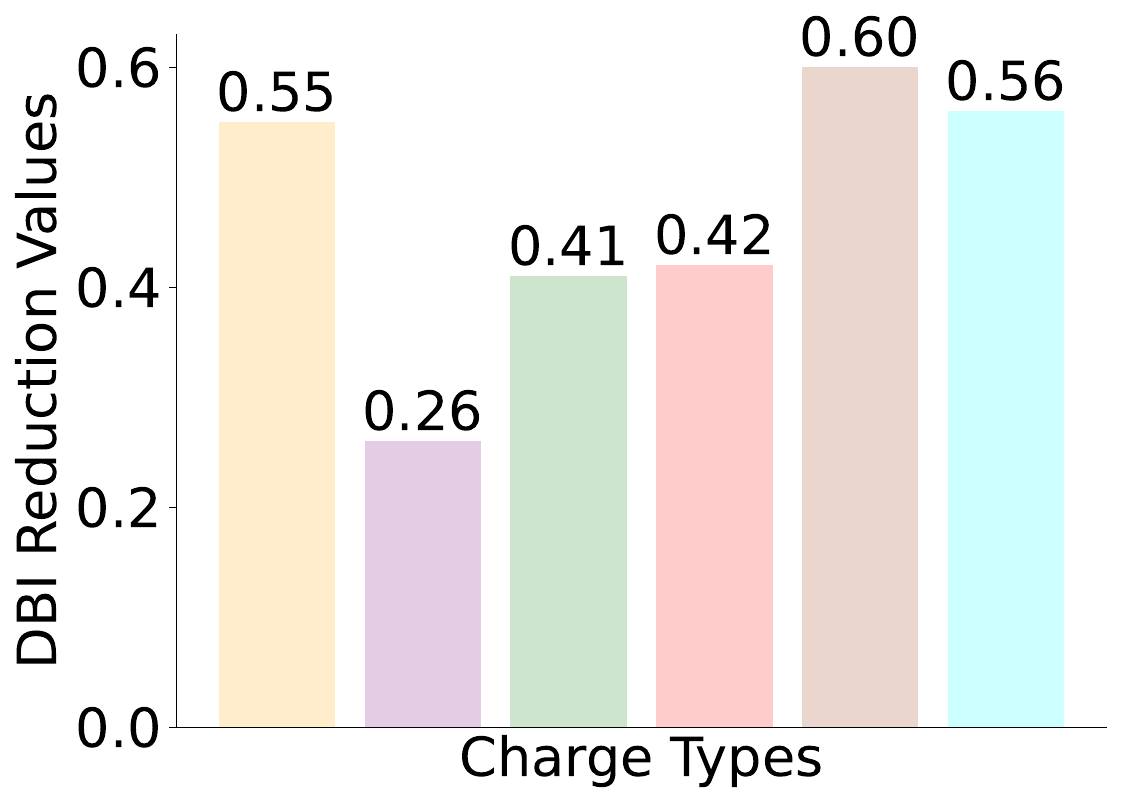}%
    }
    \hfill
    \subfigure[SAILER.\label{fig:4b}]{%
        \includegraphics[width=0.32\textwidth]{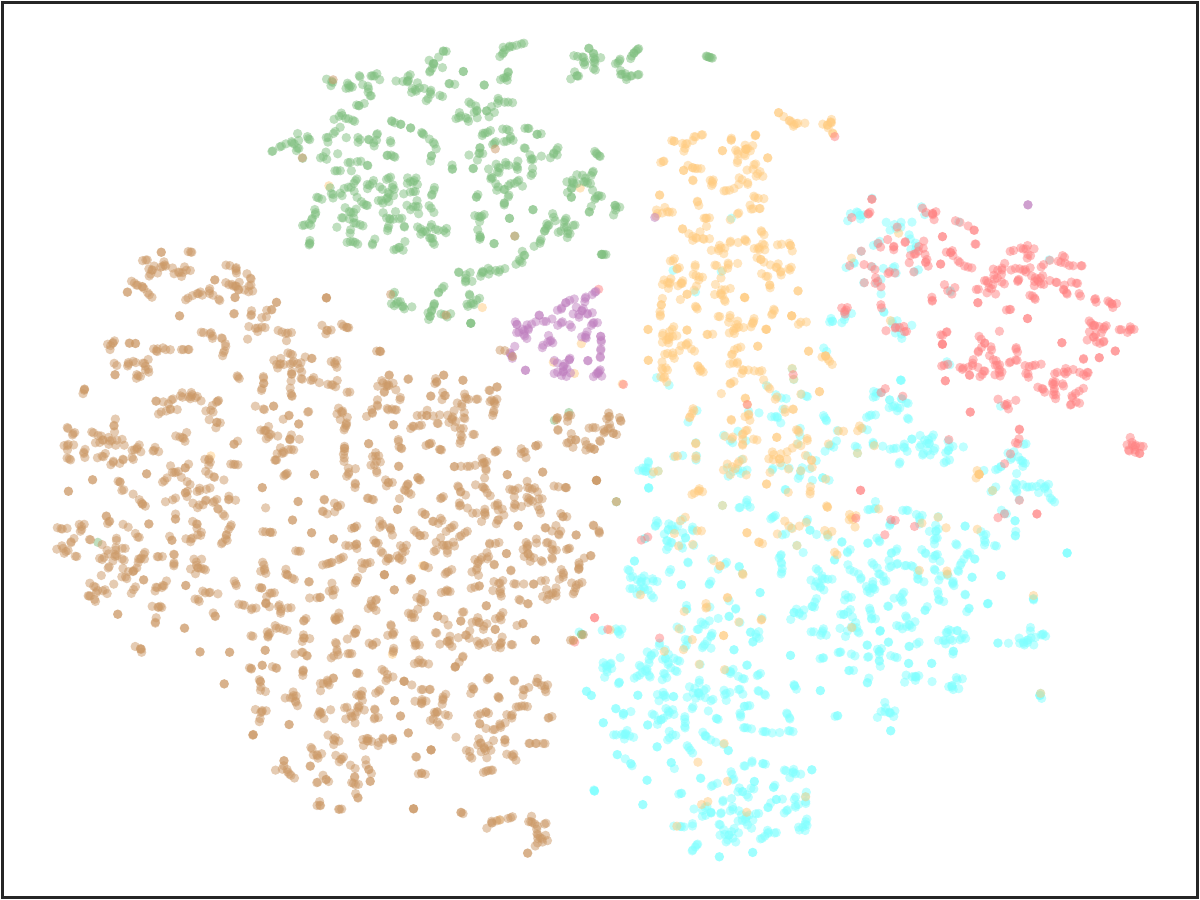}%
    }
    \hfill
    \subfigure[LegalDuet.\label{fig:4c}]{%
        \includegraphics[width=0.32\textwidth]{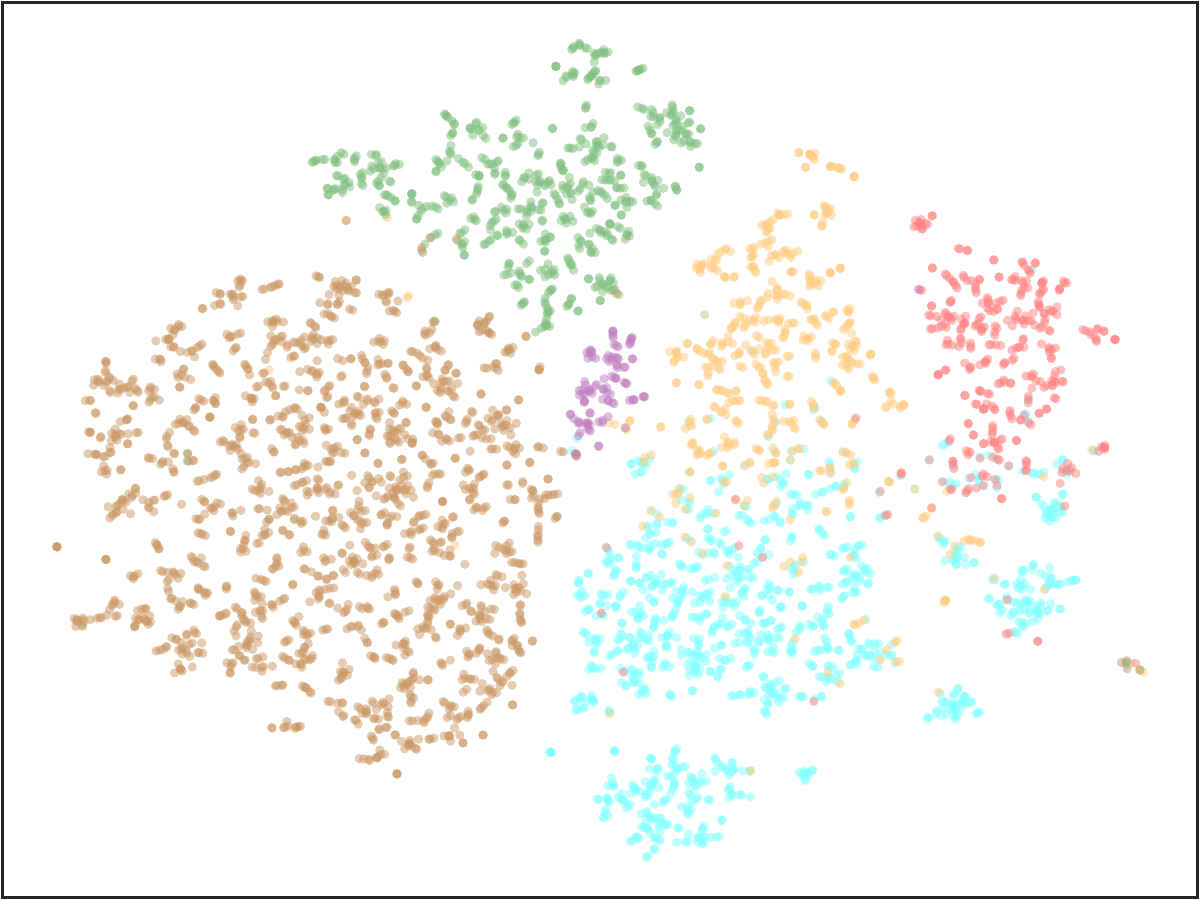}%
    }
    \caption{DBI Reduction Values and Embedding Visualizations of SAILER and LegalDuet. The embeddings of criminal facts, 
\colorbox{myorange!20}{\textbf{Provoking Troubles}},
\colorbox{mypurple!20}{\textbf{Robbery}}, 
\colorbox{mygreen!20}{\textbf{Fraud}}, 
\colorbox{red!20}{\textbf{Intentional Homicide}}, 
\colorbox{mybrown!20}{\textbf{Theft}},
\colorbox{cyan!10}{\textbf{Intentional Injury}}, 
are annotated.
}
    \label{fig:4}
\end{figure}

\textbf{Characteristics of the Learned Embedding Space by LegalDuet.}
The entropy-based analysis highlights how LegalDuet enhances prediction confidence, particularly in ambiguous cases. To further investigate its capability in fine-grained differentiation, we examine how it encodes legal cases into the embedding space, as shown in Fig.~\ref{fig:4}. 

\textit{Quantity Analyses of the Learned Embedding Space.} We first compute the Davies-Bouldin Index (DBI)~\cite{davies1979cluster}, a metric that quantifies cluster separability, where lower values indicate more compact and well-separated clusters (see Appendix~\ref{DBI} for detailed definitions of DBI). Specifically, we select six ambiguous criminal charges (Provoking Troubles, Robbery, Fraud, Intentional Homicide, Theft, and Intentional Injury) and extract all corresponding cases from the CAIL-small dataset for DBI computation. Fig.~\ref{fig:4a} illustrates the DBI reduction values when comparing the embeddings learned by SAILER and LegalDuet. The consistently positive reduction values demonstrate that LegalDuet can effectively map more legal cases into an embedding space by improving the quality of legal case clustering. Notably, LegalDuet demonstrates superior effectiveness in the tasks of Provoking Troubles, Theft, and Intentional Injury.

\textit{Visualization of Learned Embedding Space.} To provide a more intuitive view of how LegalDuet organizes criminal facts in the embedding space, we employ t-SNE to visualize the distribution of embeddings. Compared to SAILER (Fig.~\ref{fig:4b}), LegalDuet (Fig.~\ref{fig:4c}) produces a more distinct cluster behavior with clearer boundaries between similar charges, aligning with the principles of contrastive training~\cite{wang2020understanding}. This suggests that LegalDuet not only learns more structured representations but also better preserves subtle semantic distinctions between criminal facts, ultimately leading to improved classification of ambiguous charges. Additional experimental results for other pretrained language models are provided in Appendix~\ref{space}.

\section{Conclusion}
In this paper, we introduce LegalDuet, a continuous pretrained method designed to enhance language models' ability to learn more fine-grained representations for criminal facts. LegalDuet emulates the reasoning processes of judges and incorporates a dual-view legal contrastive learning mechanism. Specifically, it comprises Law Case Clustering and Legal Decision Matching, which helps PLMs to better cluster criminal facts and align the semantics between criminal facts with corresponding legal decisions. Our experiments demonstrate that LegalDuet outperforms the baseline across most evaluation metrics, instead of using more sophisticated reasoning architectures. Furthermore, its effectiveness can be generalized to various legal judgment prediction tasks and different PLMs. 




%
%
%
\bibliographystyle{splncs04}
\bibliography{mybibliography}

\clearpage

\appendix
\section{Appendix}

\subsection{Details of Legal Decision Construction}\label{Building Legal Decisions}
\begin{figure}
    \centering
    \includegraphics[width=1\linewidth]{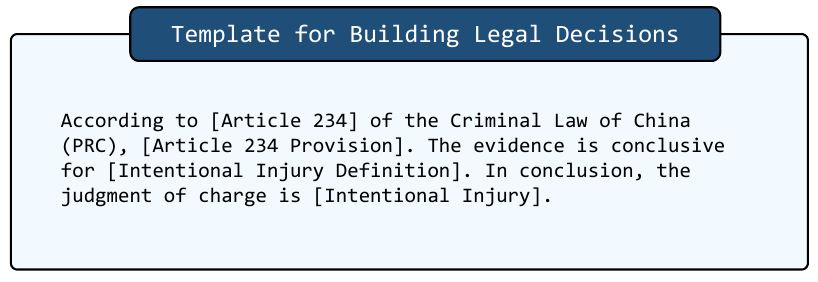}
    \caption{Legal Decision Template Used in LegalDuet.}
    \label{fig:template}
\end{figure}

We first describe the details of constructing legal decisions in Fig.~\ref{fig:template}.

\begin{table*}[htbp]
    \centering
    \caption{An Example of Legal Decision. It consists of four main parts: the law article name, the law article content, the charge definition, and the charge name.}
    \label{tab:legal_decision}
    \renewcommand{\arraystretch}{1.3} 
    \resizebox{\textwidth}{!}{ 
    \begin{tabular}{p{4cm}p{12cm}}
        \toprule
        \textbf{Component}                                  & \textbf{English Description}                                                                 \\
        \midrule
        \textbf{Law Article Name}              & Article 234                                                                       \\
        \midrule
        \textbf{Law Article Content}        & 
        Article 234 Provision: Whoever intentionally harms another person's body shall be sentenced to fixed-term imprisonment of not more than three years, criminal detention, or public surveillance. If, as a result of committing the previous crime, the victim suffers serious injury, the offender shall be sentenced to fixed-term imprisonment of no less than three years but not more than 10 years; if the victim dies, or if the offender causes serious injury leading to permanent disability through particularly cruel means, the offender shall be sentenced to fixed-term imprisonment of no less than 10 years, life imprisonment, or death penalty. If otherwise provided by law, the relevant provisions shall apply. \\
        \midrule
        \textbf{Charge Definition}   & 
        Intentional Injury Definition: Intentional Injury refers to the act of intentionally and unlawfully causing harm to another person's physical health. \\
        \midrule
        \textbf{Charge Name}         & Intentional Injury                                                               \\
        \bottomrule
    \end{tabular}
    }
\end{table*}

As shown in Table~\ref{tab:legal_decision}, our legal decision framework consists of four key components: the name of the law article ``[Article 234]'', the content of the law article ``[Provision of Article 234]'', the definition of the charge ``[Definition of Injury Definition]'', and the charge name ``[Intentional Injury]''. Logical syllogisms serve as a powerful reasoning tool, enabling rigorous analysis and sound judgment. In legal contexts, they play a crucial role by providing a structured logical framework for case analysis and adjudication.
Specifically, Article 234 establishes the major premise, forming the fundamental legal basis for judicial decision-making. It offers a foundational legal context that ensures judicial interpretation and application remain aligned with established legal principles. Additionally, the definition of a criminal charge, derived from the law article, functions as the minor premise. This distinction is essential in refining the application of the law, allowing for precise differentiation among similar criminal charges.
These provisions and definitions originate from the Criminal Law of the People's Republic of China and are instrumental in shaping legal decisions within the LegalDuet pretraining framework.


\subsection{Further Processing of Hard Negatives for Contrastive Learning}\label{Building Negative Samples}
\begin{figure*}[t]
    \centering
    \subfigure[Law Case Clustering.]{%
        \includegraphics[width=0.45\textwidth]{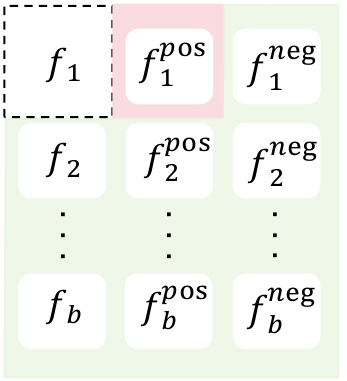}%
        \label{fig:Negative_pool_law}
    }
    \hfill
    \subfigure[Legal Decision Matching.]{%
        \includegraphics[width=0.45\textwidth]{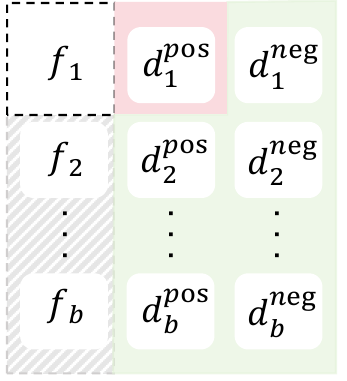}%
        \label{fig:Negative_pool_decision}
    }
   \caption{Illustration of Sample Pool Construction for Contrastive Learning. Dashed box highlights the example fact $f_1$. \colorbox{pink!30}{\textbf{Pink backgrounds}} indicate positive samples and \colorbox{mygreen!10}{\textbf{Green backgrounds}} represent negative samples. Hatched area shows other batch instances not used in this example.}
    \label{fig:Negative_pool}
\end{figure*}

As illustrated in Fig.~\ref{fig:Negative_pool}, we propose a specialized approach to leverage hard negatives for more effective contrastive learning. Our method adopts batch-aware contrastive learning, where each criminal fact $f_i$ (with $i \in \{1, 2, ..., b\}$ denoting the $i$-th instance in a batch of size $b$) is paired with positive and hard negative samples. As shown in Fig.~\ref{fig:Negative_pool}, we present an example for optimizing the representation of $f_1$ (highlighted by the dashed box), where samples with pink backgrounds represent positives, and those with green backgrounds denote negatives. The hatched area marks other batch instances that are not involved in this particular example.

\textbf{Law Case Clustering.} As shown in Fig.~\ref{fig:Negative_pool_law}, given a fact $f_i$, we first retrieve a pool of the top-15 hard negative samples using SAILER within our experimental setup. From this pool, we randomly select one criminal fact as the hard negative $f^{\text{neg}}_i$. During training, each fact $f_i$ is paired with its corresponding positive fact $f^{\text{pos}}_i$, while its complete negative sample pool comprises:
\begin{itemize}
    \item Its designated hard negative criminal fact $f^{\text{neg}}_i$.
    \item Other criminal facts in the batch ($f_j$ where $j \neq i$).
    \item All positive facts from other instances in the batch ($f^{\text{pos}}_j$ where $j \neq i$).
    \item All negative facts from other instances in the batch ($f^{\text{neg}}_j$ where $j \neq i$).
\end{itemize}

\textbf{Legal Decision Matching.} As shown in Fig.~\ref{fig:Negative_pool_decision}, for each fact $f_i$, we first construct a pool of 15 hard negative legal decisions by combining law articles and charges within our experimental setup. From this pool, we randomly sample one as the negative legal decision $d^{\text{neg}}_i$. During training, each fact $f_i$ is paired with its corresponding positive legal decision $d^{\text{pos}}_i$, while its complete negative sample pool consists of:
\begin{itemize}
    \item Its designated hard negative legal decision $d^{\text{neg}}_i$.
    \item Positive legal decisions from other instances in the batch ($d^{\text{pos}}_j$ where $j \neq i$).
    \item Negative legal decisions from other instances in the batch ($d^{\text{neg}}_j$ where $j \neq i$).
\end{itemize}

Our contrastive training strategy fully uses the instances in a batch, making these negatives maintain diversity and representative. These comprehensive negatives strengthen the ability of PLMs to learn robust and generalized representations for legal cases, allowing them to effectively distinguish between similar yet confusing cases.

\subsection{Implementation Details of Baselines}\label{implementation}
For the Feature-based baseline, we restrict the number of extracted features to the top 2,000 terms. For the CNN-based baselines, we set a maximum document length of 512 and perform word segmentation using THULAC\def\thefootnote{4}\footnote{\url{https://github.com/thunlp/THULAC-Python}}. For LSTM-based models, we retain the first 15 sentences in the criminal facts and cap each sentence at a maximum of 100 words. For PLM-based models, we set the maximum sequence length to 512.

\begin{figure}[t!]
    \centering
    \subfigure[Entropy Distribution of Charges Predictions.\label{fig:bert-entropy-accu}]{%
        \includegraphics[width=0.32\textwidth]{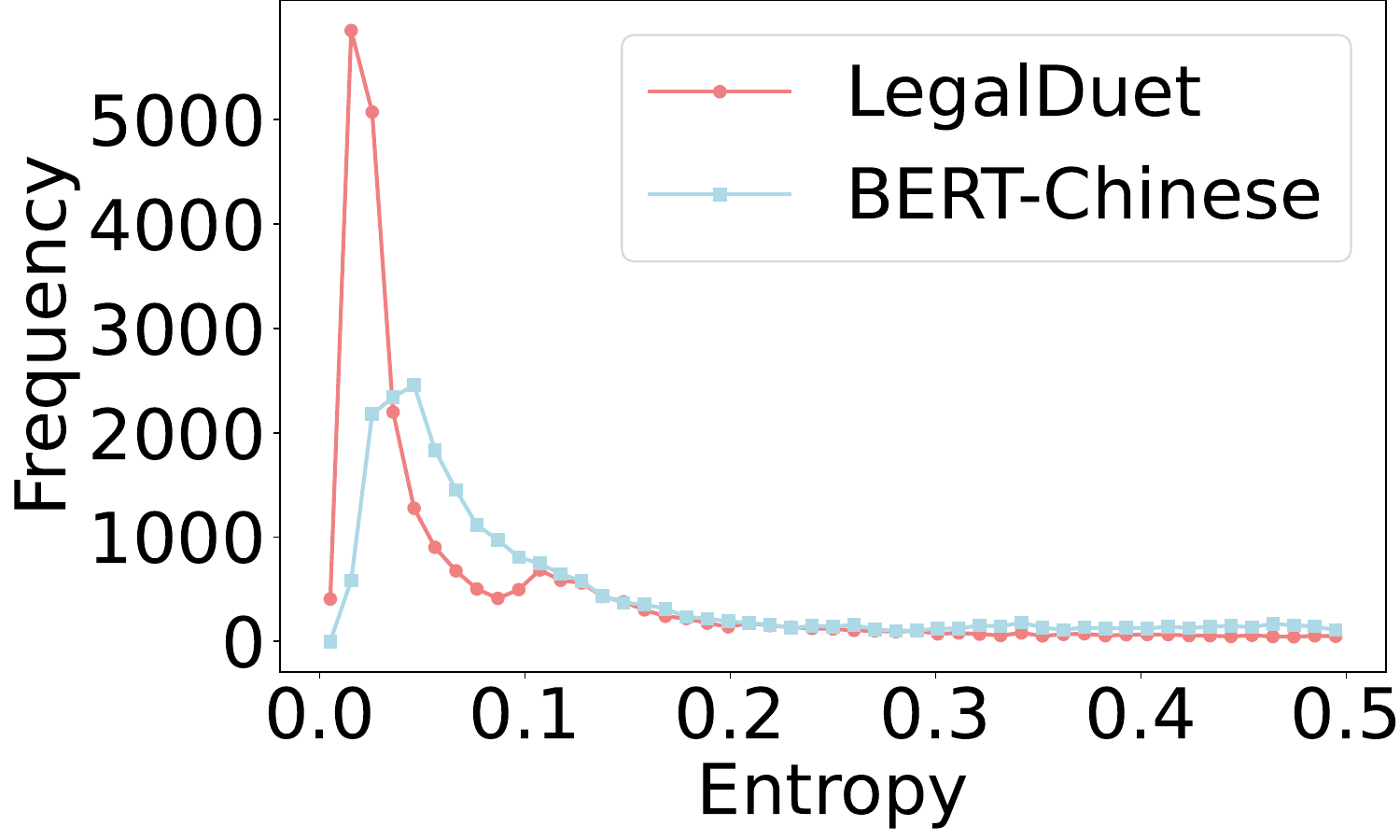}%
    }
    \hfill
    \subfigure[Entropy Distribution of Law Articles Predictions.\label{fig:bert-entropy-law}]{%
        \includegraphics[width=0.32\textwidth]{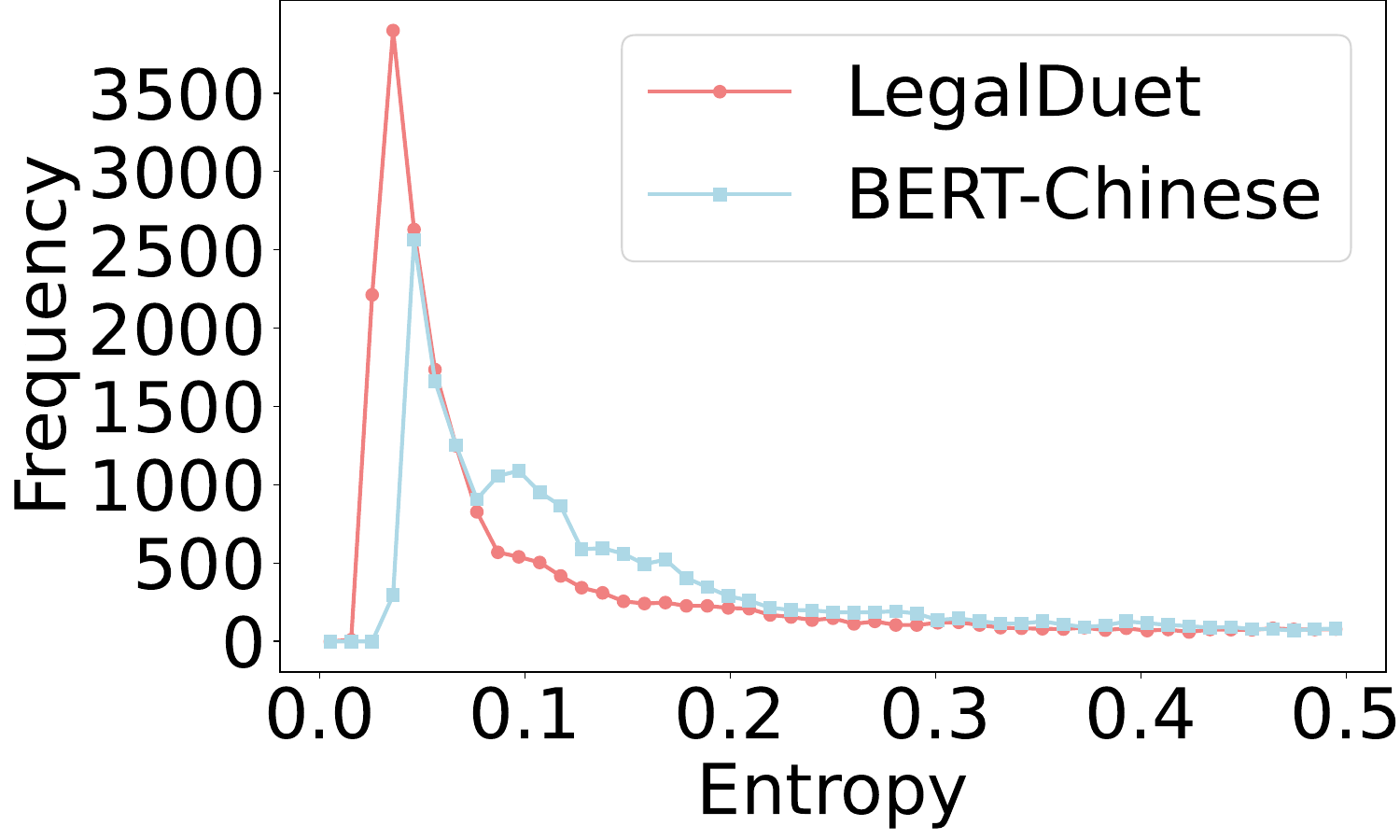}%
    }
    \hfill
    \subfigure[Entropy Distribution of Imprisonment Predictions.\label{fig:bert-entropy-term}]{%
        \includegraphics[width=0.32\textwidth]{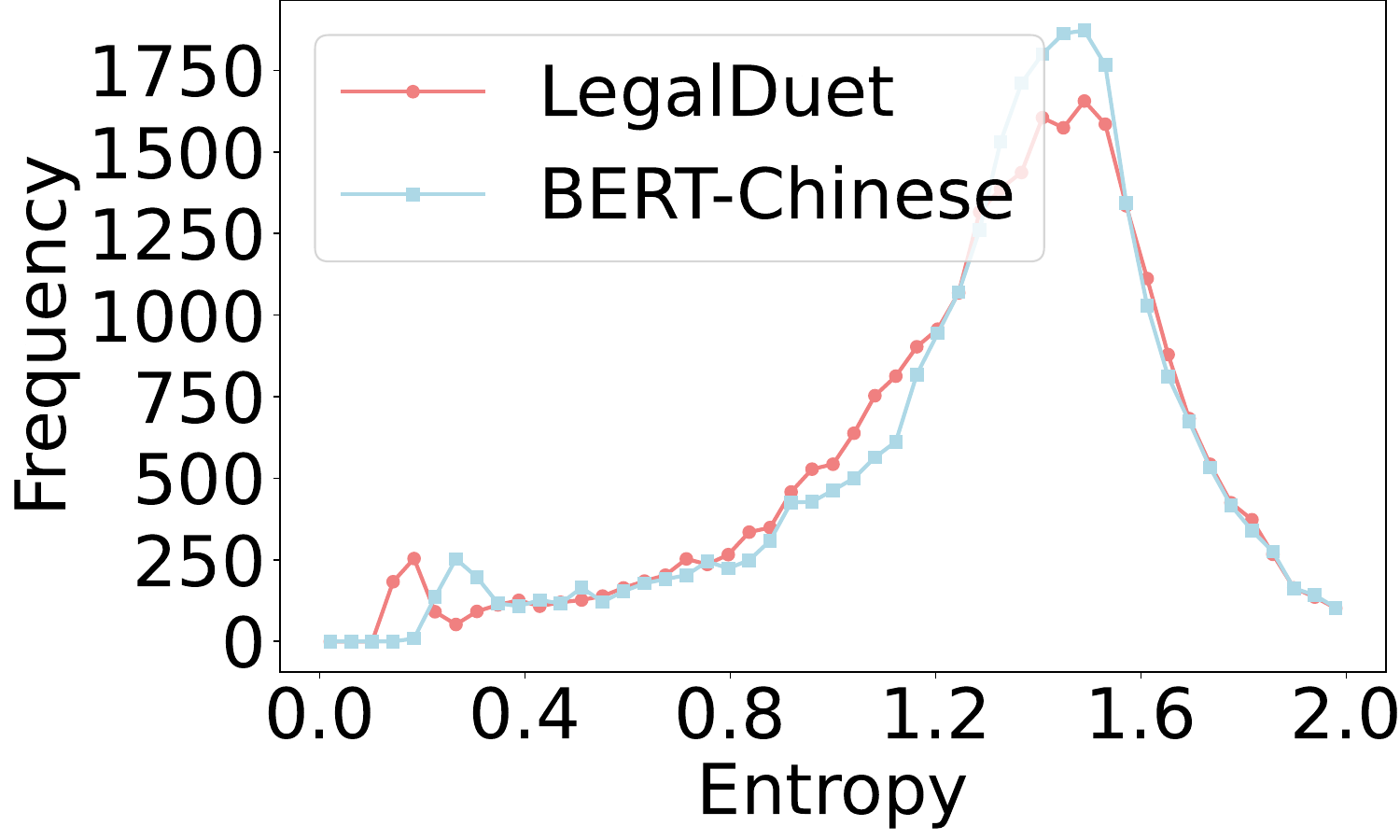}%
    }
    \hfill
    
    \caption{Entropy Distributions of Legal Judgment Predictions on BERT-Chinese.}
    \label{fig:bert-chinese-entropy}
\end{figure}

\begin{figure}[t!]
    \centering
    \subfigure[Entropy Distribution of Charges Predictions.\label{fig:bert-xs-entropy-accu}]{%
        \includegraphics[width=0.32\textwidth]{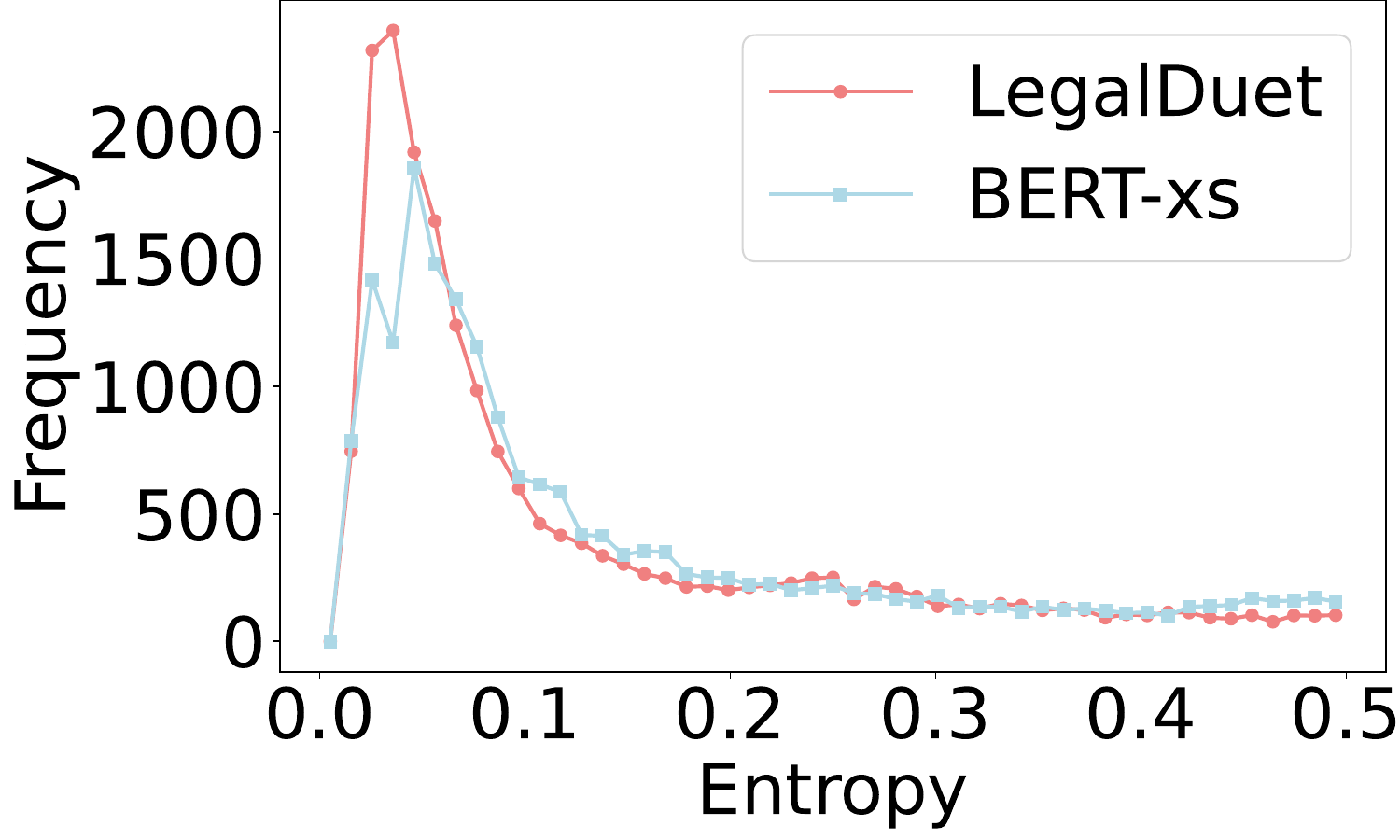}%
    }
    \hfill
    \subfigure[Entropy Distribution of Law Articles Predictions.\label{fig:bert-xs-entropy-law}]{%
        \includegraphics[width=0.32\textwidth]{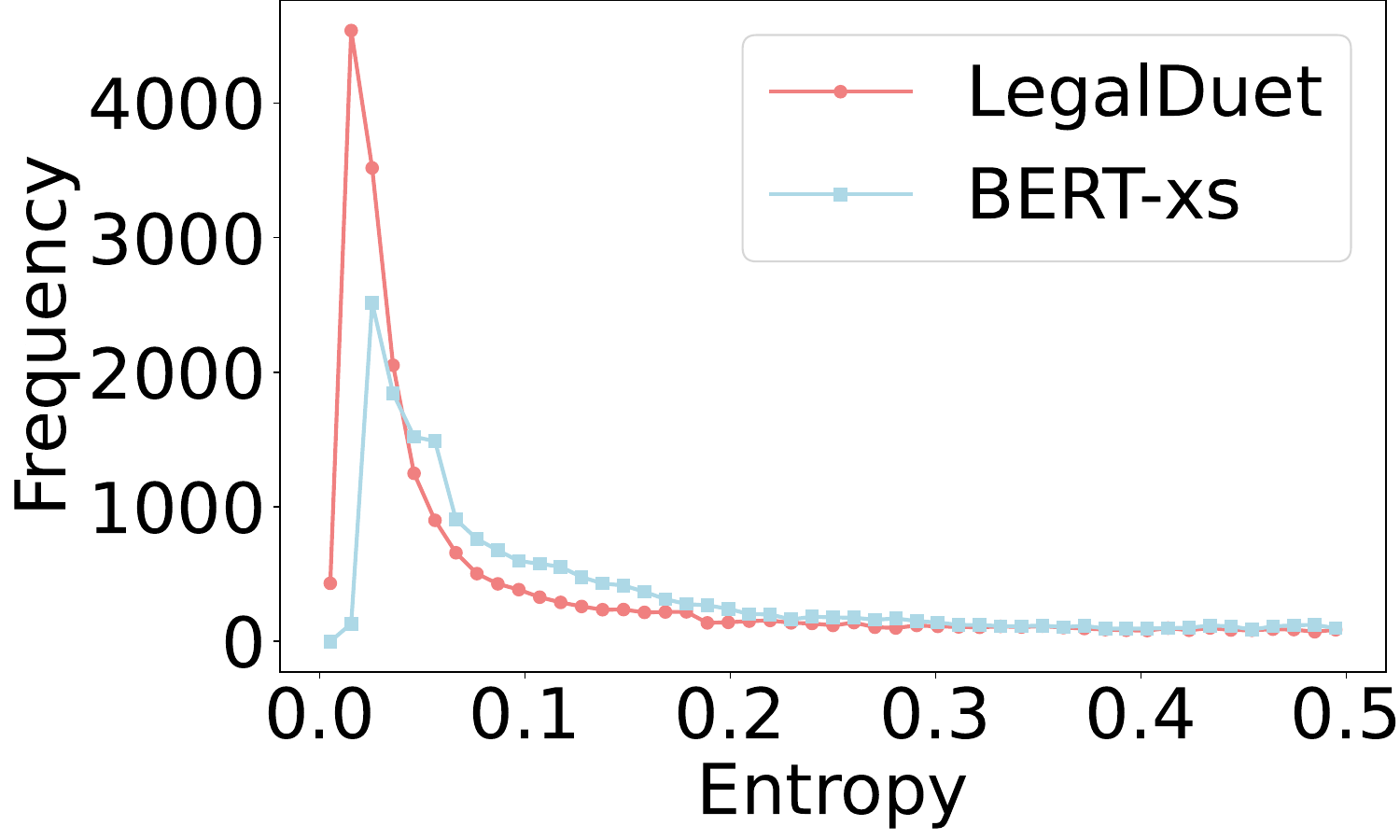}%
    }
    \hfill
    \subfigure[Entropy Distribution of Imprisonment Predictions.\label{fig:bert-xs-entropy-term}]{%
        \includegraphics[width=0.32\textwidth]{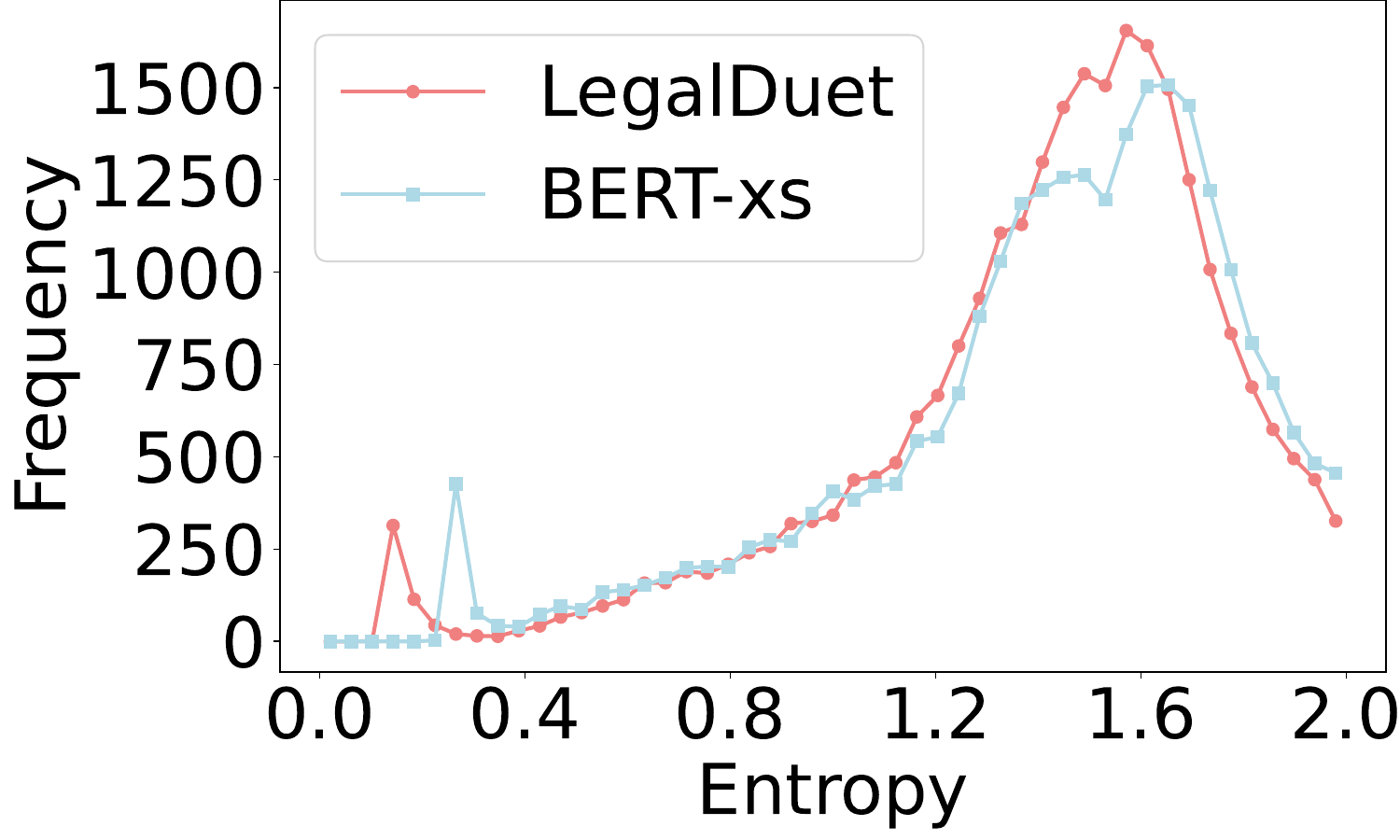}%
    }
    \hfill
    \caption{Entropy Distributions of Legal Judgment Predictions on BERT-xs.}
    \label{fig:bert-xs-entropy}
\end{figure}

\subsection{Extensive Experiment on Entropy Distribution}\label{Entropy}
We also evaluate the entropy scores for prediction tasks using both BERT-Chinese and BERT-xs, as shown in Fig.~\ref{fig:bert-chinese-entropy} and Fig.~\ref{fig:bert-xs-entropy}. The results clearly demonstrate that LegalDuet surpasses other baseline models in terms of entropy distribution performance.

\subsection{Definition of DBI Score}\label{DBI}
The Davies-Bouldin Index (DBI) is a metric used to assess the separability of the clustering results:
\begin{equation}
DBI_{i} = \max_{j \neq i} \left(  \frac{S_i + S_j} {M_{ij}} \right),
\end{equation}
where \( S_i \) is the intra-cluster distance for criminal fact related to charge \(i\) that measures the average distance between the criminal fact related to charge \(i\) and the centroid of criminal facts related to charge \(i\). \( S_j \) is the intra-cluster distance for criminal facts related to charge \(j\), defined similarly for any other charge \(j\).  \( M_{ij} \) is the inter-cluster distance between the centroids of criminal facts related to charge \(i\) and \(j\). Formally, \(S_i\) is calculated as:
\begin{equation}
S_i = \frac{1}{|C_i|} \sum_{x \in C_i} \| x - \mu_i \|,
\end{equation}
where \(C_i\) is the set of data points related to charge \(i\), \( \mu_i \) is the centroid of \(C_i\), and \(x\) represents the individual data points related to charge \(i\). Formally, \( M_{ij} \) is calculated as the Euclidean distance between the centroids \( \mu_i \) and \( \mu_j \):
\begin{equation}
M_{ij} = \| \mu_i - \mu_j \|.
\end{equation}

\subsection{Extensive Experiment on Structural Analysis and Visualization of Embedding Space}\label{space}
\begin{figure}[t]
    \centering
    \subfigure[DBI Reduction Values.\label{fig:8a}]{%
        \includegraphics[width=0.32\textwidth]{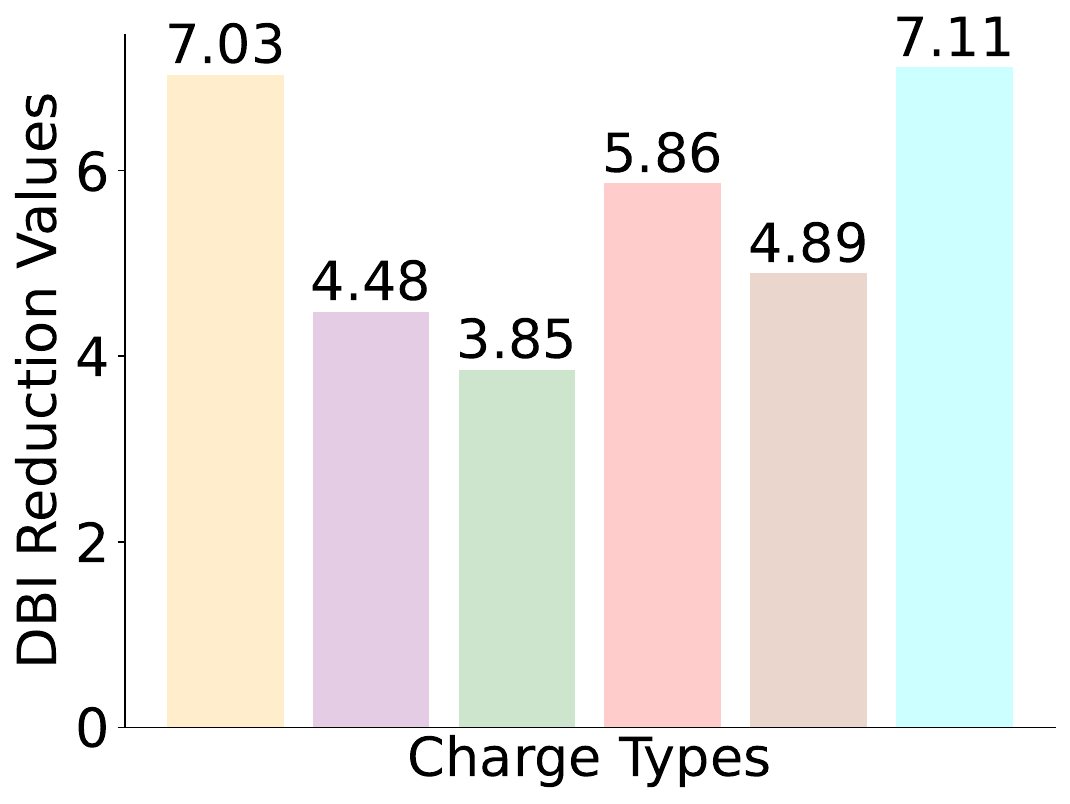}%
    }
    \hfill
    \subfigure[BERT-Chinese.\label{fig:8b}]{%
        \includegraphics[width=0.32\textwidth]{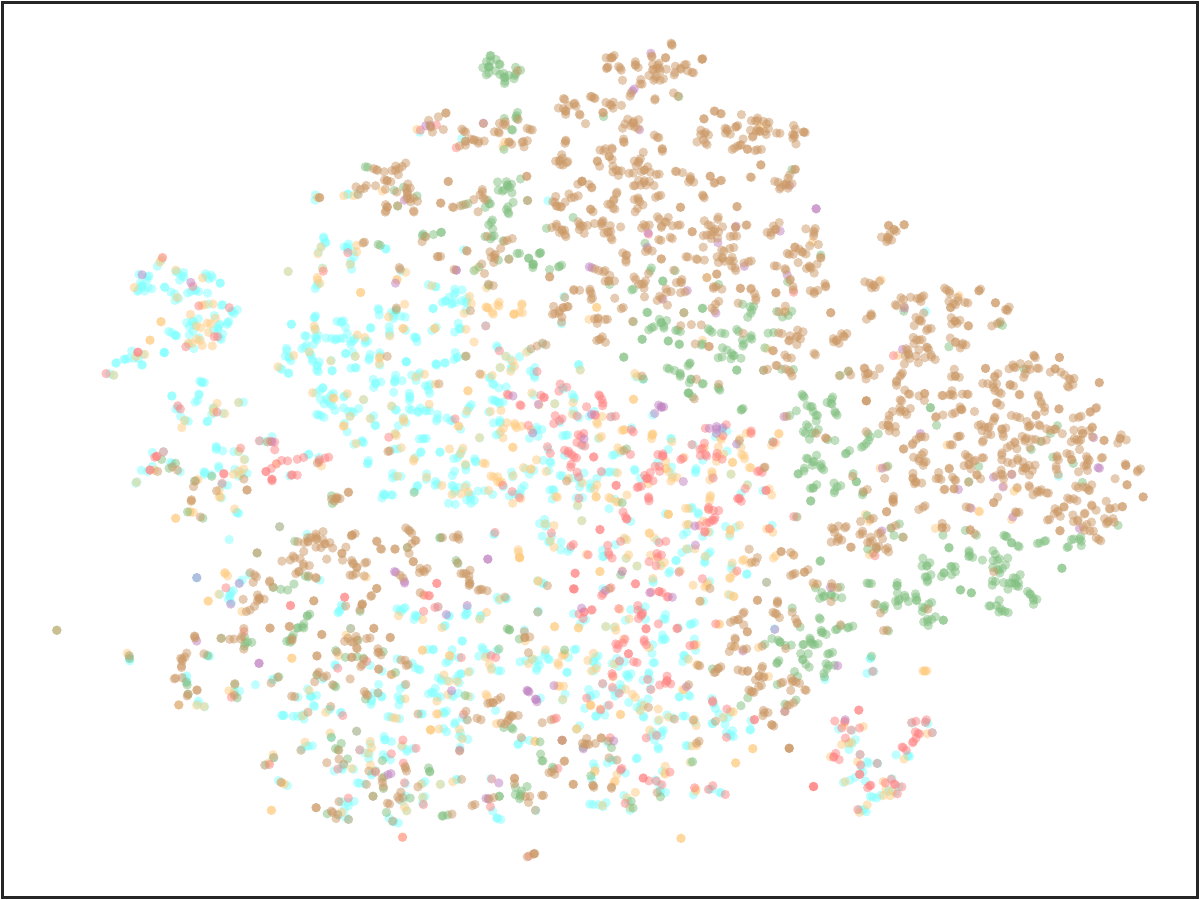}%
    }
    \hfill
    \subfigure[LegalDuet.\label{fig:8c}]{%
        \includegraphics[width=0.32\textwidth]{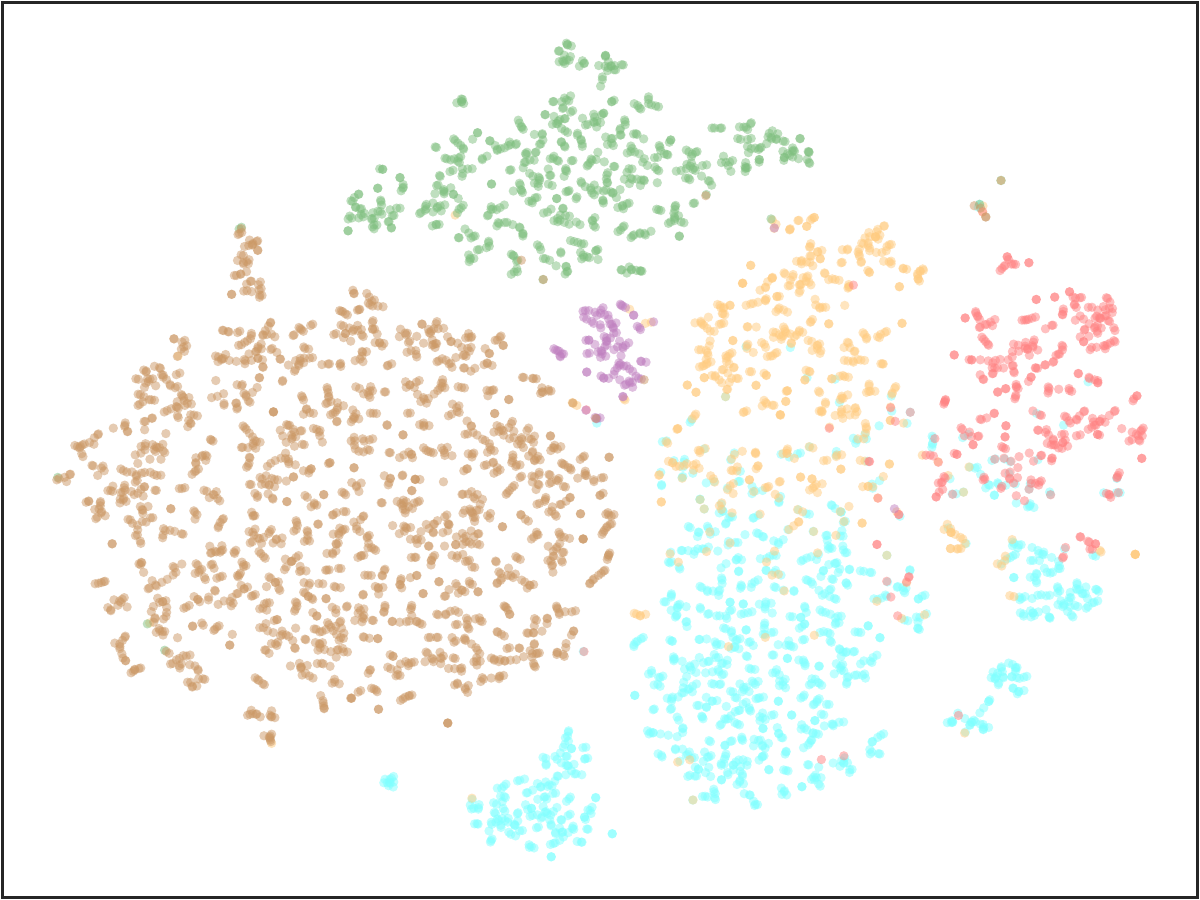}%
    }
    \caption{DBI Reduction Values and Embedding Visualizations of BERT-Chinese and LegalDuet. The embeddings of criminal facts, 
\colorbox{myorange!20}{\textbf{Provoking Troubles}},
\colorbox{mypurple!20}{\textbf{Robbery}}, 
\colorbox{mygreen!20}{\textbf{Fraud}}, 
\colorbox{red!20}{\textbf{Intentional Homicide}}, 
\colorbox{mybrown!20}{\textbf{Theft}},
\colorbox{cyan!10}{\textbf{Intentional Injury}}, 
are annotated.
}
    \label{fig:embedding_bert}
\end{figure}

\begin{figure}[t]
    \centering
    \subfigure[DBI Reduction Values.\label{fig:9a}]{%
        \includegraphics[width=0.32\textwidth]{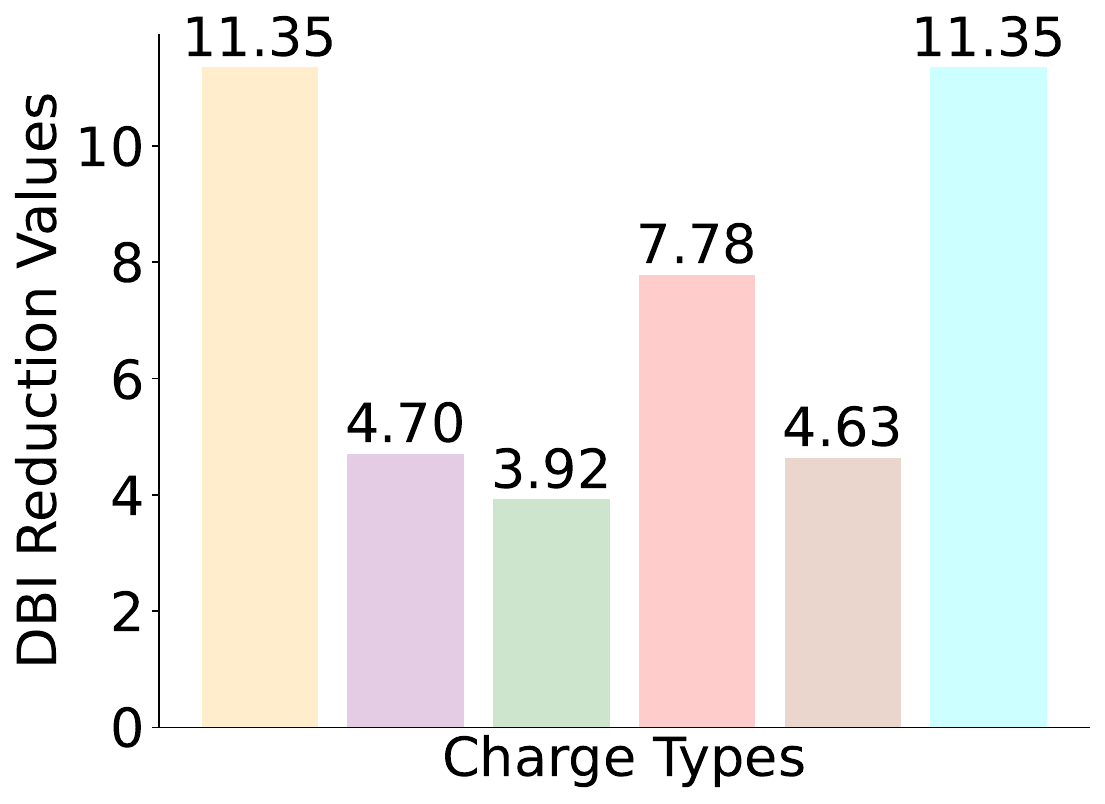}%
    }
    \hfill
    \subfigure[BERT-xs.\label{fig:9b}]{%
        \includegraphics[width=0.32\textwidth]{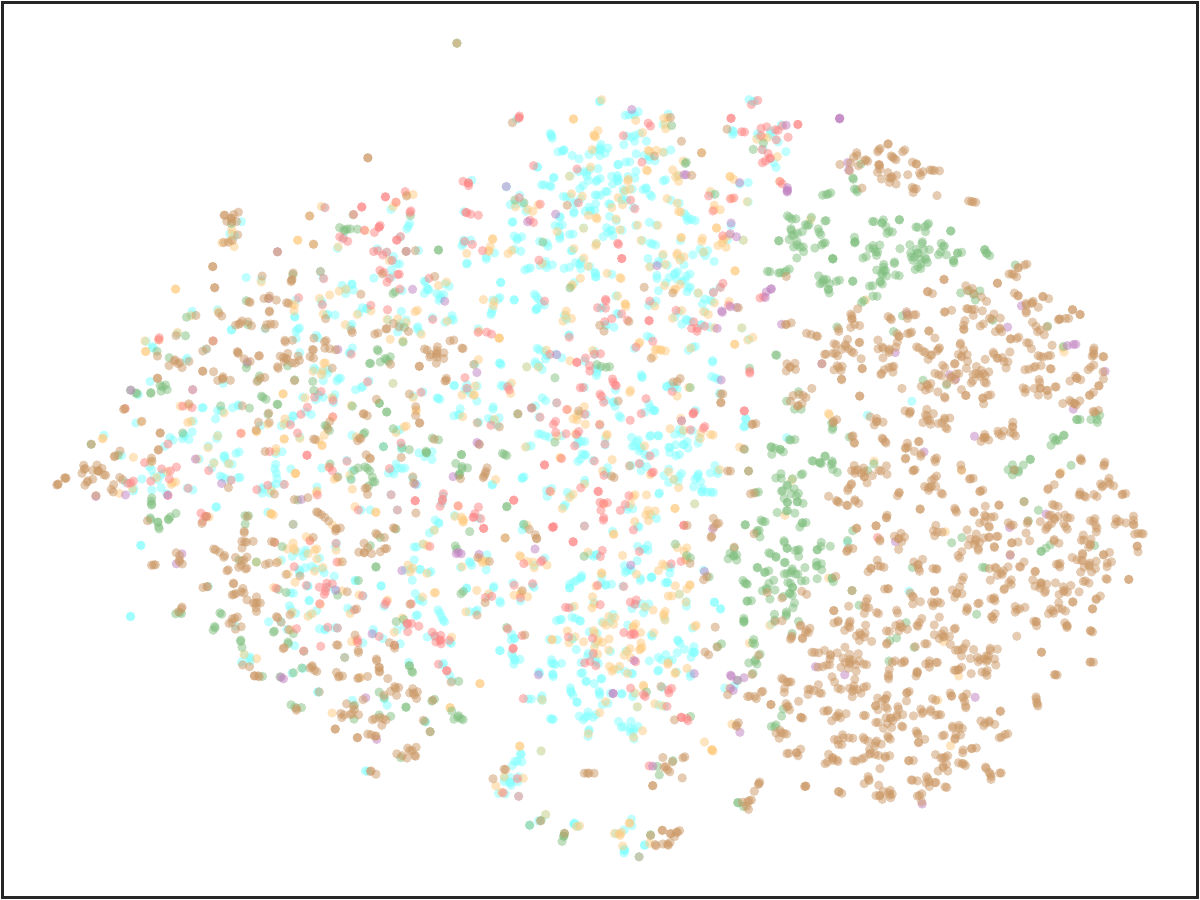}%
    }
    \hfill
    \subfigure[LegalDuet.\label{fig:9c}]{%
        \includegraphics[width=0.32\textwidth]{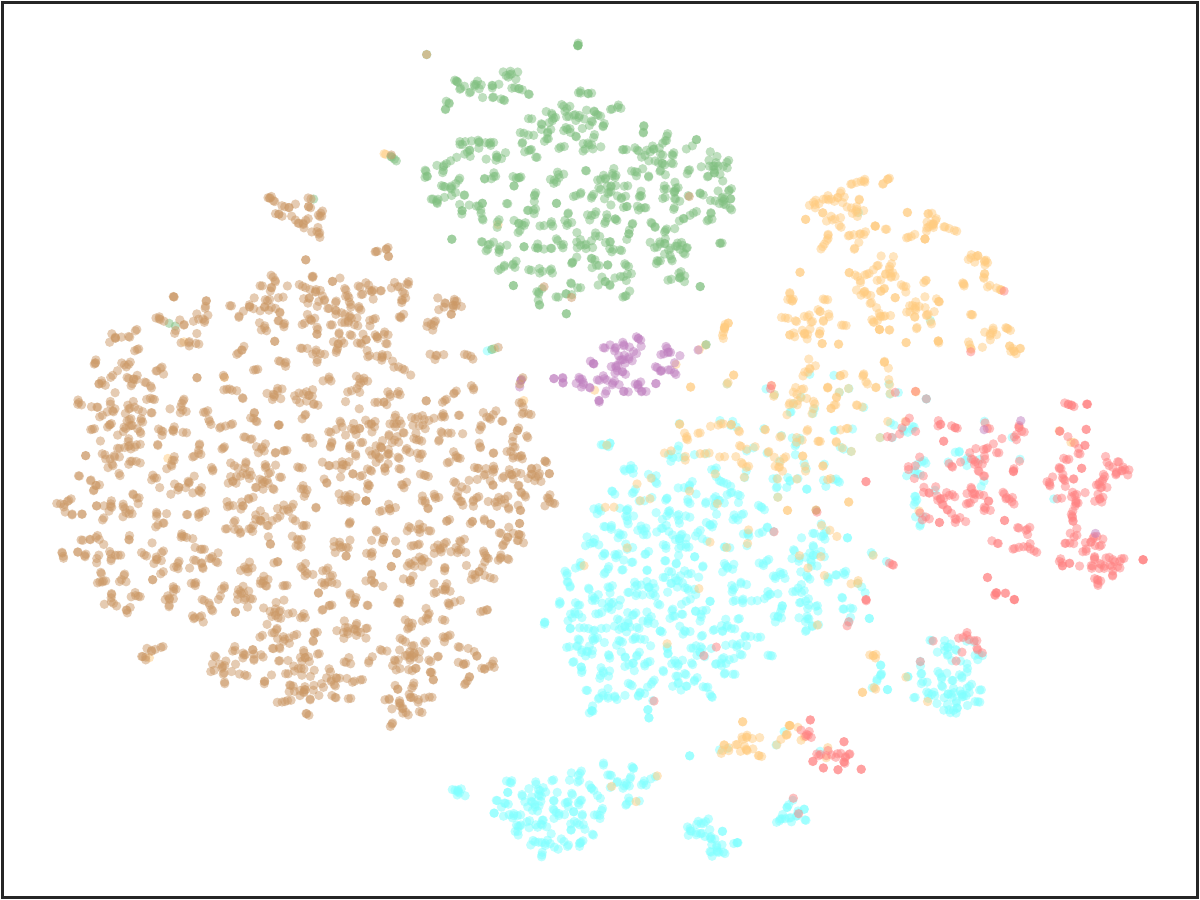}%
    }
    \caption{DBI Reduction Values and Embedding Visualizations of BERT-xs and LegalDuet. The embeddings of criminal facts, 
\colorbox{myorange!20}{\textbf{Provoking Troubles}},
\colorbox{mypurple!20}{\textbf{Robbery}}, 
\colorbox{mygreen!20}{\textbf{Fraud}}, 
\colorbox{red!20}{\textbf{Intentional Homicide}}, 
\colorbox{mybrown!20}{\textbf{Theft}},
\colorbox{cyan!10}{\textbf{Intentional Injury}}, 
are annotated.
}  
    \label{fig:embedding_bert_xs}
\end{figure}

We also compute the DBI score and visualize the embedding distributions for both BERT-Chinese and BERT-xs, as shown in Fig.~\ref{fig:embedding_bert} and Fig.~\ref{fig:embedding_bert_xs}. The results clearly demonstrate that LegalDuet achieves lower DBI scores and learns more distinct representations of criminal facts compared to other models.

\subsection{Attention Visualization of Different LJP Models}\label{attention}

\begin{figure*}[t]
    \centering
    \noindent\begin{tabularx}{\textwidth}{|X|}
    \hline \hline
    \multicolumn{1}{|c|}{
        Attention score a term takes in the passage:  
        \redhlone{1.0--1.5}
        \redhltwo{1.5--2.0} 
        \redhlthree{2.0--2.5}
        \redhlfour{2.5--3.0}
        \redhlfive{\textgreater3.0}
    } \\ \hline 

    \textcolor{blue}{Case example of SAILER:} \\
    The defendant, Qian, \redhlone{previously} worked as a \redhlone{security guard} at the \redhlone{sales office} of the XXX.
    Due to illness, he was hospitalized for several months, and later, due to dissatisfaction with the company's suggestion that he resign, he harbored grievances. On the afternoon of June 27, 2016, the defendant, Qian, \redhlone{went} to the first-floor exhibition hall of the company's sales \redhlone{office}. He \redhlone{used} a \redhlthree{lighter} to \redhlthree{ignite} a piece of paper and placed it on the model of the real estate development, causing the model to \redhlthree{catch fire}. This resulted in part of the model's buildings being \redhlone{damaged}, and the top of the model was \redhlone{blackened} by smoke. \redhlone{Fearing} the fire might \redhltwo{spread} to the \redhlone{second-floor residences}, the defendant \redhlone{used} a \redhlone{fire extinguisher} to \redhlone{put out} the \redhlthree{flames} and \redhlone{called} \redhltwo{110} to \redhlone{report the incident}. 
    \redhltwo{According to the evaluation, the economic} \redhlfour{loss} \redhltwo{caused by the} \redhlfour{fire} \redhlone{to the model of the real estate development amounted to} \redhlfour{19,800 RMB.} The prosecution alleges that the defendant, Qian, \redhlfour{deliberately caused damage} \redhltwo{to public and private} \redhlone{property} \redhlone{by} \redhlfive{setting the fire}... The facts are clear and the evidence is sufficient... \\ 
    \hline

    \textcolor{blue}{Case example of LegalDuet:} \\
    The defendant, Qian, \redhlone{previously} worked as a security guard at the sales office of the XXX.
    Due to illness, he was hospitalized for several months, and later, due to \redhlone{dissatisfaction} with the company's suggestion that he resign, he harbored grievances. On the afternoon of June 27, 2016, the defendant, Qian, went to the first-floor exhibition hall of the company's sales office. He \redhlone{used} a \redhlthree{lighter} to \redhlthree{ignite} a piece of paper and \redhlone{placed} it on the model of the real estate development, causing the model to \redhlfive{catch fire}. This resulted in part of the model's buildings being \redhlfive{damaged}, and the top of the model was \redhlone{blackened by smoke}. \redhltwo{Fearing} the \redhltwo{fire} might \redhlthree{spread} to the second-floor \redhlone{residences}, the defendant \redhlone{used} a \redhltwo{fire extinguisher} to \redhlthree{put out} the \redhlfour{flames} and \redhlone{called} \redhltwo{110} to \redhlone{report the incident}. 
    \redhlone{According to the evaluation, the} \redhltwo{economic loss caused by the} \redhlfive{fire} \redhlone{to the model of the real estate development amounted to} \redhltwo{19,800 RMB.} The prosecution alleges that the defendant, Qian, \redhlfive{deliberately caused damage}\redhlfive{to public and private property} \redhlone{by} \redhlfive{setting the fire}... The facts are clear and the evidence is sufficient... \\
    \hline \hline
    \end{tabularx}
    \caption{
        Case Study.
        In this example, SAILER predicts the charge as \textit{Arson}, while LegalDuet correctly identifies it as \textit{Intentional Damage to Property}.
        Darker red indicates higher attention scores assigned by the LJP models.
    }
    \label{fig:7}
\end{figure*}

In this section, we illustrate the contrastive learning mechanisms of SAILER and LegalDuet through attention visualization in Fig.~\ref{fig:7}. We display the attention scores assigned to different phrases.

Both SAILER and LegalDuet can allocate greater attention weights to critical phrases, such as ``setting the fire'' and ``damage'', which are crucial for enabling LJP models to make accurate legal judgment predictions. However, LegalDuet conducts more emphasis on legal clues like ``fearing'', ``spread'', and ``residence'', which indicate that the defendant Qian was concerned about the potential danger to the residents on the second floor although he set the fire. This suggests that, despite the fire causing some property damage, his actions did not directly threaten public safety, as he promptly took steps to extinguish the flames and report the incident. Additionally, LegalDuet assigns attention to the phrases, ``put out the flames'' and ``called 110'', reflecting the defendant's attempts to mitigate damage and prevent harm to public safety.

In contrast, SAILER assigns more attention to the background of the incident and the specific criminal actions, such as ``office'', ``worked as a'', ``economic loss caused by the fire'', ``RMB'', and ``setting the fire''. These phrases highlight that SAILER primarily focuses on the context and the act of arson itself, leading it to classify the crime as ``arson''. Since SAILER is trained to generate reasoning and decision components of legal cases, it tends to prioritize action-related information, particularly the defendant's role in the fire and the resulting economic damage.

The key difference lies in LegalDuet's ability to recognize the defendant's concern for public safety---evident in its attention to ``residence'' and ``spread''. This allows it to determine that the act did not pose a sufficient threat to public safety to warrant an arson charge. Instead, LegalDuet correctly identifies the crime as ``Intentional damage to property'' by assigning more attention to the legal clue ``deliberately caused damage to public and private property''. In contrast, SAILER focuses on the broader context of the act, such as ``office'', ``worked as a'', and the financial loss caused by the fire, which leads it to classify the crime as arson. By concentrating on the act itself without fully considering the defendant's intent or the actual consequences, SAILER ultimately arrives at an incorrect prediction.

\begin{table*}[t!]
\caption{Case Study on the Charge of~\textit{\textbf{Negligent Homicide}}. We present the \textcolor[rgb]{0.0, 0.2, 0.6}{prediction} and the \textcolor[rgb]{0.4, 0.0, 0.6}{Explanation} from BERT-xs, SAILER, and LegalDuet, respectively. The \textcolor[rgb]{0.7,0.3,0.3}{legal clues} with higher attention scores are highlighted.}
\resizebox{\linewidth}{!}{
\small
\begin{tabular}{p{0.20\linewidth}p{0.80\linewidth}} 
\hline
\multicolumn{2}{l}{\textbf{Case \#1 on the Charge of \textit{Negligent Homicide}}}
\tabularnewline 
\hline
\makecell[l]{Criminal Fact} & ...In 2013, Liu \textcolor[rgb]{0.698,0.298,0.302}{married} Chen A, and they had two children. Due to a conflict between Liu and Chen A, the children were sent to be \textcolor[rgb]{0.698,0.298,0.302}{fostered} at a relative's home in M County in 2014. On May 2, 2015, Liu went to visit the children at Chen A's aunt’s home in the Factory area of M County, where a \textcolor[rgb]{0.698,0.298,0.302}{dispute} broke out when Liu wanted to \textcolor[rgb]{0.698,0.298,0.302}{take the children}. After local police intervention, Liu was allowed to see the children and then \textcolor[rgb]{0.698,0.298,0.302}{left}.
On August 4, 2016, Liu traveled from X to M County and visited Yang C’s house in L Village, where Yang D (Chen A’s uncle) lived. Accompanied by his classmate Jin, Liu encountered Zhang C (Yang D’s mother) and neighbor Yang E at a shop. Liu asked Zhang C about his children, but she claimed ignorance. Noticing Chen B’s name on the shop’s business license, Liu went upstairs, followed by Zhang C, who shouted for help.
After \textcolor[rgb]{0.698,0.298,0.302}{failing to find} the children on the second floor, Liu was about to return downstairs when Zhang C \textcolor[rgb]{0.698,0.298,0.302}{locked} the anti-theft door. Liu \textcolor[rgb]{0.698,0.298,0.302}{pulled} the key from the lock and \textcolor[rgb]{0.698,0.298,0.302}{threatened} Zhang C to reveal the children’s location. Zhang C demanded the key back and called for Yang D’s help. Yang D entered the house from the shop and encountered Liu on the staircase landing, where both men grabbed each other’s collars and struggled. During the altercation, Yang D pulled Liu toward the first floor, and Liu pushed Yang D’s hand, causing Yang D to \textcolor[rgb]{0.698,0.298,0.302}{lose his balance} and \textcolor[rgb]{0.698,0.298,0.302}{fall down} the stairs, \textcolor[rgb]{0.698,0.298,0.302}{hitting} his head on the wall at the bottom...
\tabularnewline 
\hline
BERT-xs &
\textcolor[rgb]{0.698,0.298,0.302}
{\{take the children, left, failing to find, locked, pulled, threatened, hitting...\}} \newline
\textcolor[rgb]{0.0, 0.2, 0.6}{prediction}~${\Rightarrow }$~\textit{\textbf{Illegal Detention}}~(${\times}$) \\
\hline
SAILER &
\textcolor[rgb]{0.698,0.298,0.302}
{\{married, fostered, dispute, take the children, threatened, lose his balance, fall down...\}} \newline
\textcolor[rgb]{0.0, 0.2, 0.6}{prediction}~${\Rightarrow }$~\textit{\textbf{Intentional Homicide}}~(${\times}$) \\
\hline
LegalDuet &
\textcolor[rgb]{0.698,0.298,0.302}
{\{married, fostered, take the children, threatened, lose his balance, fall down, hitting...\}} \newline
\textcolor[rgb]{0.0, 0.2, 0.6}{prediction}~${\Rightarrow }$~\textit{\textbf{Negligent Homicide}}~(${\surd}$) \\
\hline
\makecell[l]{\textcolor[rgb]{0.4, 0.0, 0.6}{Explanation}} & \textit{\textbf{Illegal Detention}} refers to actions where an individual \textcolor[rgb]{0.4, 0.0, 0.6}{intentionally} and unlawfully restricts another person's \textcolor[rgb]{0.4, 0.0, 0.6}{freedom of movement}, either through \textcolor[rgb]{0.4, 0.0, 0.6}{force} or other means, without legal justification.

\textit{\textbf{Intentional Homicide}} refers to actions where an individual \textcolor[rgb]{0.4, 0.0, 0.6}{intentionally} and \textcolor[rgb]{0.4, 0.0, 0.6}{deliberately} causes the death of another person with the \textcolor[rgb]{0.4, 0.0, 0.6}{clear intent to kill}.

\textit{\textbf{Negligent Homicide}} refers to actions where an individual causes another person's death through \textcolor[rgb]{0.4, 0.0, 0.6}{carelessness} or a \textcolor[rgb]{0.4, 0.0, 0.6}{lack of reasonable caution}, without the \textcolor[rgb]{0.4, 0.0, 0.6}{intention to harm}.
\tabularnewline 
\hline
\end{tabular}
}
\label{tab:new_case_study_1}
\end{table*}

\begin{table*}[t!]
\caption{Case Study on the Charge of~\textit{\textbf{Provoking Troubles}}. We present the \textcolor[rgb]{0.0, 0.2, 0.6}{prediction} and the \textcolor[rgb]{0.4, 0.0, 0.6}{Explanation} from BERT-xs, SAILER, and LegalDuet, respectively. The \textcolor[rgb]{0.7,0.3,0.3}{legal clues} with higher attention scores are highlighted.}
\resizebox{\linewidth}{!}{
\small
\begin{tabular}{p{0.20\linewidth}p{0.80\linewidth}} 
\hline
\multicolumn{2}{l}{\textbf{Case \#2 on the Charge of \textit{Provoking Troubles}}}
\tabularnewline 
\hline
\makecell[l]{Criminal Fact} &{...Wang disputed with Lin, a ride-hailing driver, over a ride issue. The defendant, Wang, immediately opened the front passenger door of the vehicle driven by Lin and \textcolor[rgb]{0.698,0.298,0.302}{entered} the passenger seat. He verbally abused Lin and prepared to \textcolor[rgb]{0.698,0.298,0.302}{physically assault} him. Subsequently, the defendant took out a \textcolor[rgb]{0.698,0.298,0.302}{spring knife} that he had on him and \textcolor[rgb]{0.698,0.298,0.302}{threatened} the victim, Lin. During a \textcolor[rgb]{0.698,0.298,0.302}{struggle} for the spring knife, Lin sustained a cut on his finger. The defendant then \textcolor[rgb]{0.698,0.298,0.302} {forced} the victim to \textcolor[rgb]{0.698,0.298,0.302} {hand over 110 RMB} in cash. After receiving the money, the defendant \textcolor[rgb]{0.698,0.298,0.302}{left} the scene...}
\tabularnewline 
\hline
BERT-xs
&
\makecell[l]{\textcolor[rgb]{0.698,0.298,0.302} {\{entered, spring knife, struggle, hand over, RMB...\}}\\\textcolor[rgb]{0.0, 0.2, 0.6}{prediction}~${\Rightarrow }$~\textit{\textbf{Robbery}} (${\times}$)}
\tabularnewline 
\hline 
\makecell[l]{SAILER}
&
\makecell[l]{\textcolor[rgb]{0.698,0.298,0.302} {\{struggle, forced, hand over, RMB, left...\}}\\\textcolor[rgb]{0.0, 0.2, 0.6}{prediction}~${\Rightarrow }$~\textit{\textbf{Blackmail}}(${\times}$)}
\tabularnewline 
\hline 
LegalDuet  
&
\makecell[l]{\textcolor[rgb]{0.698,0.298,0.302} {\{physically assault, threatened, forced, hand over, 110 RMB...\}}\\\textcolor[rgb]{0.0, 0.2, 0.6}{prediction}~${\Rightarrow }$~\textit{\textbf{Provoking Troubles}} (${\surd}$)}
\tabularnewline 
\hline 
\makecell[l]{\textcolor[rgb]{0.4, 0.0, 0.6}{Explanation}}& {\textit{\textbf{Robbery}} refers to actions where an individual unlawfully takes another person's \textcolor[rgb]{0.4, 0.0, 0.6}{property} through the use of force or intimidation, with the intent to \textcolor[rgb]{0.4, 0.0, 0.6}{permanently} deprive the victim of their possessions. 

\textit{\textbf{Blackmail}} refers to the act of unlawfully obtaining money or property from a victim with the intent of illegal possession, through means of \textcolor[rgb]{0.4, 0.0, 0.6}{coercion}, \textcolor[rgb]{0.4, 0.0, 0.6}{intimidation}, or \textcolor[rgb]{0.4, 0.0, 0.6}{threats of harm}.  

\textit{\textbf{Provoking Troubles}} refers to actions where an individual disrupts public order without the specific intent to acquire property or coerce someone, often involving aggressive or unruly behavior aimed at \textcolor[rgb]{0.4, 0.0, 0.6}{creating disturbances} and \textcolor[rgb]{0.4, 0.0, 0.6}{asserting dominance} rather than obtaining financial benefits.} 
\tabularnewline 
\hline
\end{tabular}
}
\label{tab:new_case_study_2}
\end{table*}
\subsection{Case Study}\label{case}

In this subsection, we present two criminal facts with confusing charges \textit{Negligent Homicide} and \textit{Provoking Troubles} to illustrate the effectiveness of LegalDuet in Table~\ref{tab:new_case_study_1} and Table~\ref{tab:new_case_study_2}. We compare LegalDuet against legal domain PLMs, specifically BERT-xs and SAILER, to highlight its superior ability in capturing subtle legal distinctions.

In the first case, both BERT-xs and SAILER misclassify the charge, whereas LegalDuet correctly predicts \textit{Negligent Homicide}. LegalDuet identifies key details such as ``hitting'' and ``losing his balance'', correctly determining that the victim's injury resulted from an accidental fall rather than intentional violence. This aligns with the legal definition of \textit{Negligent Homicide}, which involves causing harm due to carelessness or lack of caution, without intent to kill. While LegalDuet also recognizes contextual elements like ``threatened'' and ``take the children'', its emphasis on the accidental nature of the injury ensures the correct classification.
In contrast, BERT-xs overemphasizes terms, such as ``take the children'', ``failing to find'', ``pulled'', and ``locked'', leading it to erroneously associate these actions with \textit{Illegal Detention}. However, these terms are more accurately interpreted as obstructive or interfering behaviors during the conflict rather than deliberate attempts to unlawfully restrict another person's freedom. Similarly, SAILER places excessive focus on the term ``dispute'' and interprets the physical struggle as an escalating conflict, ultimately misclassifying the case as \textit{Intentional Homicide}. This misclassification arises from overlooking the crucial distinction that the injury was not the result of deliberate violence.

In the second case, LegalDuet correctly classifies the charge as \textit{Provoking Troubles} by identifying legal cues related to both the criminal action and the nature of harm, such as  ``physically assault'', ``threatened'', ``forced'', ``hand over'', and ``110 RMB''. These terms indicate disruptive and aggressive behavior without the intent for significant financial gain or severe violence, aligning well with the legal definition of \textit{Provoking Troubles}.
In contrast, BERT-xs emphasizes phrases like ``entered'', ``spring knife'', and ``struggle'', leading it to misclassify the charge as \textit{Robbery} by overemphasizing physical confrontation and weapon use---elements that do not meet the threshold for the severe coercion required under \textit{Robbery}. Similarly, SAILER fixates on terms like ``struggle'', ``forced'', and ``RMB'', incorrectly associating them with \textit{Blackmail} by emphasizing coercion and financial elements while failing to recognize the absence of threats of exposure or reputational damage, which are essential components of \textit{Blackmail}.

%




\end{document}